\pdfoutput=1	

\documentclass[journal]{IEEEtran}
\usepackage{epsfig,graphicx}
\usepackage{amsthm,multirow,cite,balance}
\usepackage[cmex10]{amsmath}
\usepackage{color}
\usepackage[small]{caption}
\usepackage{epstopdf}
\usepackage{array}
\usepackage{rotating}
\usepackage{varwidth}
\usepackage{booktabs}
\usepackage{amssymb}
\usepackage{algorithm}
\usepackage{algpseudocode}
\usepackage{mathtools}
\usepackage{pdfpages}
\usepackage{footnote}
\usepackage{placeins}
\usepackage{verbatim} 
\usepackage{bm}
\usepackage{mathtools}
\usepackage{subcaption}
\usepackage{enumitem}
\usepackage{booktabs}
\usepackage{multirow}
\usepackage{siunitx}
\usepackage{empheq}
\usepackage{enumitem}
\usepackage{url}
\setlist{parsep=0pt,listparindent=\parindent}

\DeclarePairedDelimiter\floor{\lfloor}{\rfloor}

\def\L{{\cal L}}

\newcommand{\ba}{{\bf a}}

\newcommand{\by}{{\bf y}}

\newcommand{\bw}{{\bf w}}

\newcommand{\bW}{{\bf W}}

\newcommand{\bT}{{\bf T}}

\DeclareMathOperator*{\argmin}{argmin}

\ifCLASSINFOpdf
\else
\fi
\begin{document}
\title{Robust and Low-Rank Representation \\for Fast Face Identification with Occlusions }

\author{Michael~Iliadis,~\IEEEmembership{Student~Member,~IEEE,}        
        %Albert~S.~Berahas,~\IEEEmembership{Student~Member,~IEEE,}
	   Haohong~Wang,~\IEEEmembership{Member,~IEEE,}
	   Rafael~Molina,~\IEEEmembership{Member,~IEEE,}
        and~Aggelos~K.~Katsaggelos,~\IEEEmembership{Fellow,~IEEE}% <-this % stops a space

\thanks{This paper has been partially supported by the Spanish Ministry of Economy and Competitiveness under project TIN2013-43880-R, the European Regional Development Fund (FEDER).}
\thanks{M.~Iliadis and~A.~K.~Katsaggelos are with the Department of EECS, Northwestern University, Evanston, IL 60208-3118 USA (e-mail: miliad@northwestern.edu; aggk@eecs.northwestern.edu).}
%\thanks{A.~Berahas is with the Department of Engineering Sciences and Applied Mathematics, Northwestern University, Evanston, IL 60208-3118 USA (e-mail: albertberahas2012@u.northwestern.edu ).}
\thanks{H.~Wang is with TCL Research America, San Jose, CA 95131 USA (e-mail: haohong.wang@tcl.com).}
\thanks{R.~Molina is with the Departamento de Ciencias de la Computaci\'on e I.A., Universidad de Granada, 18071 Granada, Spain (e-mail: rms@decsai.ugr.es).}}

\maketitle

\begin{abstract}

In this paper we propose an iterative method to address the face identification problem with block occlusions. Our approach utilizes a robust representation based on two characteristics in order to model contiguous errors (e.g., block occlusion) effectively. The first fits to the errors a distribution described by a tailored loss function. The second describes the error image as having a specific structure (resulting in low-rank in comparison to image size). We will show that this joint characterization is effective for describing errors with spatial continuity. Our approach is computationally efficient due to the utilization of the Alternating Direction Method of Multipliers (ADMM). A special case of our fast iterative algorithm leads to the robust representation method which is normally used to handle non-contiguous errors (e.g., pixel corruption). Extensive results on representative face databases (in constrained and unconstrained environments) document the effectiveness of our method over existing robust representation methods with respect to both identification rates and computational time.

Code is available at Github, where you can find implementations of the F-LR-IRNNLS and F-IRNNLS (fast version of the RRC \cite{Yang2013}) : \url{https://github.com/miliadis/FIRC}  

\end{abstract}


% \fi
\IEEEpeerreviewmaketitle

\section{Introduction}\label{sec:introduction}

\begin{figure}[!t]
     \centering
	\includegraphics[scale=0.35]{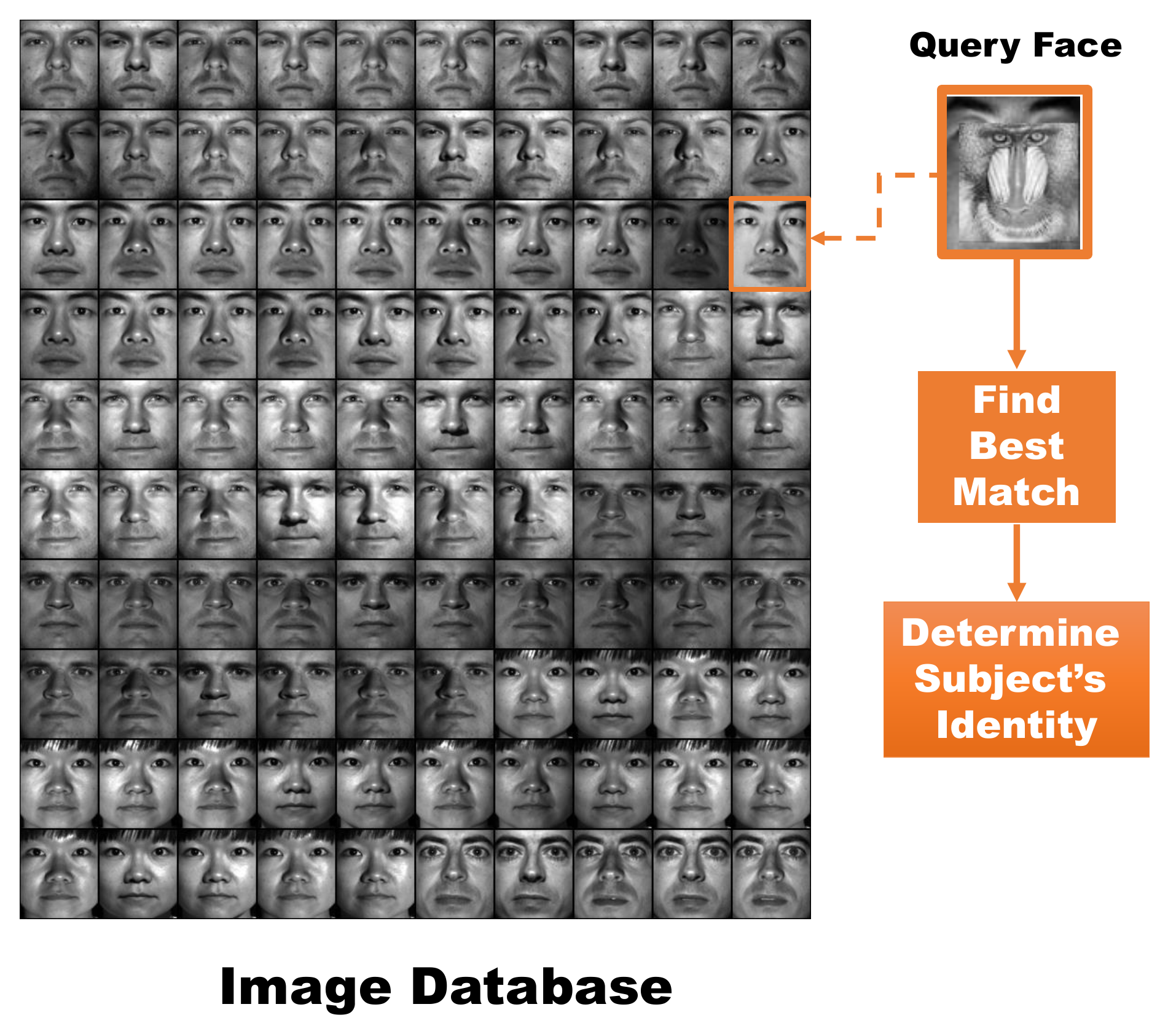}\\
     %\vspace{-0.3cm}
    \caption{Overview of the face identification problem investigated in this work. The image database consists of non-occluded faces of different subjects with small illumination changes while the query face exhibits 70\% block occlusion that covers most of the informative features of the face.}
     %\vspace{-0.8cm}
     \label{fig:first_page}
\vspace{-0.7cm}
\end{figure}

\IEEEPARstart{F}{ace Identification} (FI) focuses on deducing a subject's identity through a provided test image and is one of the most popular problems in computer vision \cite{Zhao2003,Taigman2015}. Typically, test images exhibit large variations, such as occlusions. Ideally, if the training set contains the same type of occlusion as the test image then identification becomes a rather straightforward task. In practice, however, there is no guarantee that the collected data would cover all different occlusions for all identities of interest. An example of this problem is presented in Figure~\ref{fig:first_page}. The image database consists of non-occluded faces of subjects with intra-class illumination differences while the query face exhibits a 70\% random block occlusion that covers most of the informative features of the face. In applications where prior knowledge such as the region and the object of occlusion is not provided, an appropriate modeling of the error between the test image and the training samples is necessary. 

%Recent advances in computer vision have led to improved results in image recognition by utilizing deep learning representations \cite{Krizhevsky2012}. In faces, deep learning focuses on learning face representations \cite{Taigman2014,Sun2014,Sun2014a,Sun2015,Taigman2015,Schroff2015,Sun2015a} that are robust to unconstrained environments where illumination, pose, expression and exposure is uncontrolled. While a lot of progress has been made \cite{Huang2007}, a large number of training samples is required to achieve high performance \cite{Sun2015a,Taigman2015,Hu2015}. What is more, when the critical eye regions are occluded the performance degrades significantly with over 50\% occlusion \cite{Sun2015a}.

Early works on face identification \cite{Turk1991,Belhumeur1997} attempted to deal with illumination variations. The concept of $\ell_1$-graph, which is robust to data noises and naturally sparse, was introduced in \cite{Cheng2010} to encode the overall behavior of the training set in sparse representations. To handle more complex variations such as face disguise and expressions, sparse representation-based classification models were proposed \cite{Wright2009,Wright2010,Wagner2012,Zhang2011,Shi2011}. The main idea in these approaches is that a subject's face sample can be represented as a linear combination of available images of the same subject. Then, the face class that yields the minimum reconstruction error is selected in order to classify the subject. One recent extension of the sparse representation-based classification model is the class-wise sparse representation \cite{Lai2016}. In this method, the number of training classes is minimized to alleviate the problem of representing the query by samples from many different subjects. Another extension is the patch-based classification approach \cite{Lai2016,Lai2012}. The patch based approach employs the sparse representation-based classifier to each patch of the face separately and the final decision is made by fusing the patch classification results. Notice that, in patch-based approaches the way to partition the image might be critical for the identification performance, especially when occlusion affects all patches \cite{Lai2012}.

\begin{figure*}[!t]
     \centering
	\includegraphics[scale=0.67]{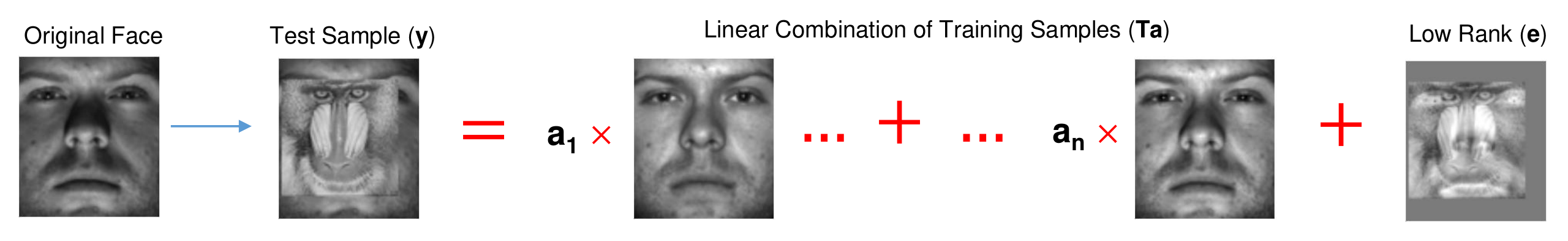}\\
     %\vspace{-0.3cm}
    \caption{Degradation Model: the test sample with occlusion can be represented as the linear combination of training samples with some intra-class variations (e.g., lighting) plus the error ${\bf e}$. The error ${\bf e}$ has two characteristics; it is considered low-rank in comparison of image size and follows a distribution described by a tailored potential loss function.}
     \vspace{-0.5cm}
     \label{fig:error_description}
\end{figure*}

To address cases with complex occlusion and corruption robust representation models\cite{Wright2009,Yang2011,Yang2013,He2014,He2011,He2013,Lai2016} of the error image\footnote{Error image is the difference between the occluded test face and the unoccluded training face of the same identity.} were considered, utilizing a non-Gaussian distribution model to minimize the influence of outliers. In these models a Laplacian distribution (sparse error) or more general distributions based on M-Estimators \cite{Huber1981,He2014} were fitted to the errors. There are, however, two main drawbacks with such approaches. First, the iterative reweighted algorithm used to solve the robust representation problem is computationally expensive when dealing with high-dimensional data \cite{Yang2011,Yang2013,He2014,He2011}. Second, the performance degrades with over 50\% block occlusion. According to \cite{Li2013}, this is due to the assumption that error pixels are independent. The robust representation approaches are usually effective in FI cases with non-contiguous\footnote{We call variations such as block occlusion and face disguise (e.g., scarves) {contiguous errors} since the error image is zero everywhere except in the region of the occluded object.} errors such as pixel corruption.

In cases of contiguous errors there is a spatial correlation among the error pixels as mentioned in \cite{Zhou2009,Li2013,Qian2014}. To exploit this correlation, the spatial continuity of the error image was integrated into the sparse representation-based classification model \cite{Zhou2009,Li2013}. However, these models lack convergence guarantees\cite{Qian2014}. To address this issue Qian et. al. \cite{Qian2014} observed that the error image with contiguous occlusion is low-rank and proposed to estimate the error support by solving a nuclear norm minimization problem with the use of ADMM \cite{Boyd2011}. While the low-rank assumption was well justified the method was effective with up to 50\% random block occlusion. A reason might be that only the structure of the error (error support) was exploited and not its distribution (e.g., sparsity). 

Low-rank estimation has been considered in \cite{Ding2015} where a discriminative low-rank metric learning method was proposed that jointly learns a low-rank linear transformation matrix and a low-rank representation. Authors in \cite{Li2013a} developed a graph construction model, with robust similarity metric (low-rank representation, which is robust to noisy data) and balanced property for the application of semi-supervised learning. In \cite{Li2014}, a dictionary learning algorithm with low-rank regularization for FI was proposed with Fisher discriminant function to the coding coefficients to make the dictionary more discriminative.

Given corrupted training samples, one can utilize robust PCA (RPCA) \cite{Cand`es2011} or its variants, such as, supervised low-rank (SLR) \cite{Jiang2015} and double nuclear norm-based matrix decomposition (DNMD) \cite{Zhang2015} to recover the ``clean" data. Since these methods target to recover the original data matrix, they are transductive. Unlike these approaches, inductive methods aim at learning an underlying projection matrix from training data to remove possible corruptions in a new datum. Two popular inductive methods are the inductive RPCA (IRPCA) \cite{Bao2012} and inductive DNMD (IDNMD) \cite{Zhang2015}, respectively. In the experiments conducted in \cite{Bao2012,Zhang2015} same occlusion and corruption was utilized in both training and testing images (e.g., in \cite{Zhang2015} the same occlusion object was used in both training and testing samples). The limitation of these approaches is that the corruption to be handled in the new datum should be similar to corruption present in the training data.

In this work, we propose an iterative method to solve the FI problem with occlusions. We consider the same scenario as in \cite{Yang2011,Yang2013,Li2013} and \cite{Qian2014} according to which we are given ``clean" frontal aligned views with a block occlusion which appear in any position on the test image but is ``unseen" to the training data. When corrupted training data are provided that are not frontally aligned (e.g., scenarios in an unconstrained environment) a tool such as RPCA and SLR is employed to separate outlier pixels and corruptions from the training samples as a pre-processing step. Then, the ``clean" frontal aligned faces are used for training data to perform face identification with occluded test images. 

As already mentioned, with the robust methods \cite{Yang2011,Yang2013,Qian2014,Li2013} high computational cost is exhibited and identification results significantly degrade with over 50\% random block occlusion. Our approach is based on a new iterative method which is efficient in terms of computational cost and robust to block occlusions up to 70\%. To describe contiguous errors (e.g., a random block occlusion) the proposed method utilizes a robust representation with two characteristics. The first fits to the errors a distribution based on a tailored loss function. The second models the reduced rank structure of the error in comparison to image size. We will then show that this joint characterization is effective for describing errors with spatial continuity. Our approach is efficient in terms of computational cost. Efficient minimization is performed by reformulating the reweighted coding problem as a constrained ADMM one, thus, avoiding costly matrix inversions. 

A special case of our fast iterative algorithm leads to the robust representation method presented in \cite{Yang2011,Yang2013} which is normally used to handle non-contiguous errors (e.g., pixel corruption).

The rest of the paper is organized as follows. In Section~\ref{sec:firc} first we present our fast iterative algorithm for FI cases with occlusion. Then, we describe the identification scheme and the weight function used in our method. Finally, a special case of the algorithm for solving efficiently the FI problem with non-contiguous errors is described. Experimental results and discussion on the performance of the proposed algorithms are presented in Section~\ref{sec:exps}, and conclusions are drawn in Section~\ref{sec:conclusions}.

\section{Robust Face Identification with Block Occlusions}\label{sec:firc}

In this section we propose an iterative method to address the FI problem with block occlusions. Our approach utilizes the robust representation of \cite{Yang2011,Yang2013,He2014,He2011} with two characteristics and uses a tailored loss function based on M-estimators. The method handles contiguous errors that are considered low-rank in comparison to the size of the image and is efficient in terms of computational cost. A special case of our method is also utilized to solve efficiently the robust representation problem for non-contiguous errors.

Let ${\bf y} \in \mathbb{R}^d$ denote the face test sample in a column-wise vectorized form where $j \times k = d$ is the size of the image. Let ${\bf T} = [ T_i, \ldots, T_c] \in \mathbb{R}^{d \times n}$ denote the matrix (dictionary) with the set of samples of $c$ subjects stacked in columns. ${T_i} \in \mathbb{R}^{d \times n_i}$ denotes the $n_i$ set of samples of the $i^{th}$ subject, such that, $\sum_i n_i = n$. 

As illustrated in Figure~\ref{fig:error_description} we can represent the test sample with occlusion as the superposition of training samples and a representation error ${\bf e}$, thus, the degradation model is, 
\begin{align}
{\bf y} = {\bf Ta} + {\bf e},
\label{eq:deg_model}
\end{align}
where ${\bf e} \in \mathbb{R}^d$ is the representation error and ${\bf a} \in \mathbb{R}^n$ is the representation vector. Thus, the test sample can be represented as a linear combination of the samples in ${\bf T}$. The face identity is chosen based on the minimum class residual provided by the estimated coefficients ${\bf a}$ as in \cite{Wright2009}. The residual image ${\bf e}$ in our model has two characteristics:
\begin{enumerate}
  \item It includes mainly the occluded object, and therefore it is considered low-rank in comparison to image size since many of its rows or columns are zero. Please note, that we will be using the term ``low-rank" from here onward in this paper to indicate errors that have reduced rank in comparison to the size of the image.
  \item It follows a distribution that can be effectively described by a tailored potential loss function (e.g., the logistic function utilized in \cite{Yang2013}).
\end{enumerate} 
We expect that the calculation of ${\bf e}$ based on the two characteristics mentioned above will lead to an accurate estimation of the ${\bf a}$ coefficients and provide the correct identity. Although the first characteristic has been employed in \cite{Qian2014} and the second in robust representation methods such as \cite{Yang2013}, both of them are necessary to adequately describe the residual image and are used together for the first time in our work. In particular, enforcing only the second characteristic will not necessarily lead to an error image that is structured and low-rank. To the best of our knowledge there is no study that utilizes both the robust representation and the structure of the error using low-rank estimation in a unified framework. 

For the problem described above, we propose the following function to be minimized,
\begin{equation}
\label{eq:philr}
J({\bf a}) = \sum\limits_{i=1}^d\phi(({\bf y- \bf T\bf a})_i) + \lambda_*\left\Vert T_M({\bf y}-{\bf Ta})\right\Vert _* + \vartheta{\left({\bf a}\right)},
\end{equation}
where  $\phi: \mathbb{R} \rightarrow \mathbb{R}$ is a potential loss function which is selected from a variety of  M-Estimators \cite{Huber1981,He2014} and $({\by} - {\bT\ba})_i$ is the $i^{th}$ component error. The function $\vartheta({\bf a})$ is used as a regularizer of the coefficients ${\bf a}$, $\lambda_*>0$ and $T_M$ is an operator that transforms its vector argument into a matrix of appropriate size such that $T_M({\bf y}-{\bf Ta}) \in \mathbb{R}^{j \times k}$. The nuclear norm $\left\Vert\cdot\right\Vert _*$ is the convex-relaxation of the rank function. 

Thus we are looking in minimizing the following problem,
\begin{equation}
\label{eq:min_ja}
\min_{\bf a} J({\bf a}).
\end{equation}
Notice, how previous works relate to our model:
\begin{enumerate}
\item For $\phi(x) = x^2$, $\lambda_* = 0$ and $\vartheta({\bf a}) = \lambda||{\bf a}||_1$ with $\lambda >0$, it is the sparse representation-based classification (SRC) \cite{Wright2009} given as,
\begin{equation}
\label{eq:l1-min}
\min_{\bf a}\left\Vert {\bf y}-{\bf Ta}\right\Vert _{2}^{2} + \lambda||{\bf a}||_1.
\end{equation}
\item For $\phi(x) = x^2$ and $\vartheta({\bf a}) = \lambda||{\bf a}||_2^2$, it is the low-rank regularized regression ($\text{LR}^3$) \cite{Qian2014} which is formulated as,
\begin{equation}
\label{eq:lowrank-min}
\min_{\bf a}\left\Vert {\bf y}-{\bf Ta}\right\Vert _{2}^{2} + \lambda_*\left\Vert T_M({\bf y}-{\bf Ta})\right\Vert _* + \lambda\left\Vert {\bf a}\right\Vert _{2}^{2}.
\end{equation} 
\item For $\lambda_* = 0$, it is the robust representation problem \cite{Yang2011,Yang2013,He2014,He2011} formulated as,
\begin{equation}
 \min_{\bf a}  \sum\limits_{i=1}^d{\phi(({\by} - {\bT\ba})_i)} + \vartheta{\left({\bf a}\right)}.
\label{eq:ja}
\end{equation}
\end{enumerate}

In previous works authors have chosen different functions $\vartheta({\bf a})$ to regularize the coefficients ${\bf a}$. In the collaborative representation-based classification with regularized least square (CR-RLS) \cite{Zhang2011} the authors are solving the SRC problem with $\vartheta({\bf a}) = \lambda||{\bf a}||_2^2$. In \cite{Yang2011, He2014} $\vartheta({\bf a}) = \lambda||{\bf a}||_1$ was used combined with different potential functions while in the regularized robust coding (RRC) \cite{Yang2013}, $\vartheta({\bf a}) = \lambda||{\bf a}||^2_2$ was used. In correntropy-based sparse representation (CESR) \cite{He2011} and structured sparse error coding (SSEC) \cite{Li2013}, $\vartheta({\bf a})$ was chosen to be the indicator function of the nonnegative orthant $\mathbb{R}^n_+$, such that a nonnegative ${\bf a} \ge 0$ regularization term was enforced.    

In this work we choose $\vartheta({\bf a})$ so as the representation coefficients are nonnegative, since it has been shown to provide robust representation for FI in \cite{He2011} and \cite{Li2013}. 

The robustness property of $\phi(\cdot)$ in \eqref{eq:philr}, to be described now, in combination with $\vartheta(\cdot)$ will force some of the coefficients of $\bf e$ to be zero or very small in magnitude. It will also force $\bf a$ to be concentrated in areas of the images that can be represented well by faces in $\bT$. Furthermore, the use of the nuclear norm will force the residual to be low-rank.

We consider potential loss functions $\phi(\cdot)$ symmetric around zero associated to Super Gaussian (SG) distributions\cite{Babacan2012}.
The function $\phi(\sqrt{x})$ has to be increasing and concave for $x \in (0,\infty)$ \cite{Palmer2010}. This condition is equivalent to $\phi'(x)/x$ being decreasing on $(0,\infty)$, that is, for $x_1\ge x_2 \ge 0$,  $\phi'(x_1)/x_1 \le \phi'(x_2)/x_2$. If this condition is satisfied, then $\phi(\cdot)$ can be represented as (using \cite[Ch. 12]{Rockafellar:1996}),
\begin{align}
\phi\left(x\right) &= \inf_{\xi>0} \,\frac{1}{2} \, \xi \, x^2 \, -\phi^*\left(\frac{1}{2}\xi\right),
\label{eq:infxfg}
\end{align}
where $\phi^*\left(\xi\right)$ is the concave conjugate of $\phi(\sqrt{x})$ and $\xi$ is a variational parameters. The dual relationship to (\ref{eq:infxfg}) is given by \cite{Rockafellar:1996},
\begin{align}
\phi^*\left(\frac{1}{2}\xi\right) =  \inf_{x} \,\frac{1}{2} \, \xi \, x^2- \phi(x) \, .\label{eq:dual}
\end{align}
Equality in (\ref{eq:infxfg}) is obtained at the optimal values of $\xi$, which are computed from the dual representation (\ref{eq:dual}) by taking the derivative with respect to $x$ and setting it to zero, which gives $\xi = \phi'(x)/x$. By using \eqref{eq:infxfg} we can write the function in \eqref{eq:philr} as, 
\begin{equation}
    \begin{aligned}    
\label{eq:hqrrclr}
{J}({\bf {a}}) = &\min_{\bf w}\frac{1}{2}\Arrowvert \sqrt{\bf W}({\bf y}-{\bf Ta})\Arrowvert _{2}^{2} + {\varphi({\bf w})} \\ 
&+ \lambda_*\left\Vert T_M({\bf y}-{\bf Ta})\right\Vert _* + \vartheta{\left({\bf a}\right)},
 \end{aligned}
\end{equation}
where $\bw=(\xi_1,\ldots, \xi_d)$ with $\xi_i>0, i=1,\ldots, d$, $\bW=\mbox{diag}(\bw)$, and $\varphi(\bw)=\sum\limits_{i=1}^d\phi^*(\frac{1}{2}\xi_i)$. 

Notice, before proceeding, that the weights in $\bf{W}$ are of the form $\phi'(x)/x$ which are large for small values of $x$ for SG potential loss functions, so $\bf a$ will fit well small values of $\bf e$ in magnitude.

Let us consider the augmented function by subsistuting \eqref{eq:hqrrclr} into \eqref{eq:min_ja},
\begin{equation}
    \begin{aligned}    
\label{eq:aug}
{J}({\bf {a}},{\bf {w}}) = &\frac{1}{2}\Arrowvert \sqrt{\bf W}({\bf y}-{\bf Ta})\Arrowvert _{2}^{2} + {\varphi({\bf w})} \\ 
&+ \lambda_*\left\Vert T_M({\bf y}-{\bf Ta})\right\Vert _* + \vartheta{\left({\bf a}\right)}.
 \end{aligned}
\end{equation}
A local minimizer $({{\bf a},{\bf w}})$ can be calculated by alternating minimization in two steps; in step one, the weights are updated by fixing the representation coefficients ${\bf a}$ and in step two the vector ${\bf a}$ is updated by fixing the weights in ${\bf W}$, i.e.,
\begin{subequations}
\label{eq:irls_second}
\begin{align}
\label{eq:weight_lr_step}
{w_i^{t+1}} &= \phi'(({\by- \bT\ba^t})_i)/({\by- \bT\ba^t})_i\\
\label{eq:coding_lr_step}
{\bf {a}}^{t+1} &= \argmin_{\bf a}\Arrowvert \sqrt{{\bf W}^{t+1}}({\bf y}-{\bf Ta})\Arrowvert_{2}^{2} \nonumber \\ 
&+ \lambda_*\left\Vert T_M({\bf y}-{\bf Ta})\right\Vert _* + \vartheta{\left({\bf a}\right)},
\end{align}
\end{subequations}
where ${w_i^{t+1}}$ and  ${{\bf a}}^{t+1}$ are the weights and representation coefficients estimated at the $t^{th}$ iteration, respectively. The term $({\bf y} - {\bf Ta}^t)_i $ denotes the component error $i$ at the $t^{th}$ iteration. Large weights are assigned to pixels in the residual image with small errors in the previous reconstructed iteration, while small weights are assigned to pixels in the residual image with large errors. 

We expect the use of the nuclear norm of the residual to force small weights in {\bf W} to be assigned only on the occluded part. Thus, outlier pixels will not contribute much to the reconstruction at the next iteration, and at convergence, the estimated error will mainly consist of those outliers. 

\begin{figure*}[!t]
     \centering
	\includegraphics[scale=0.50]{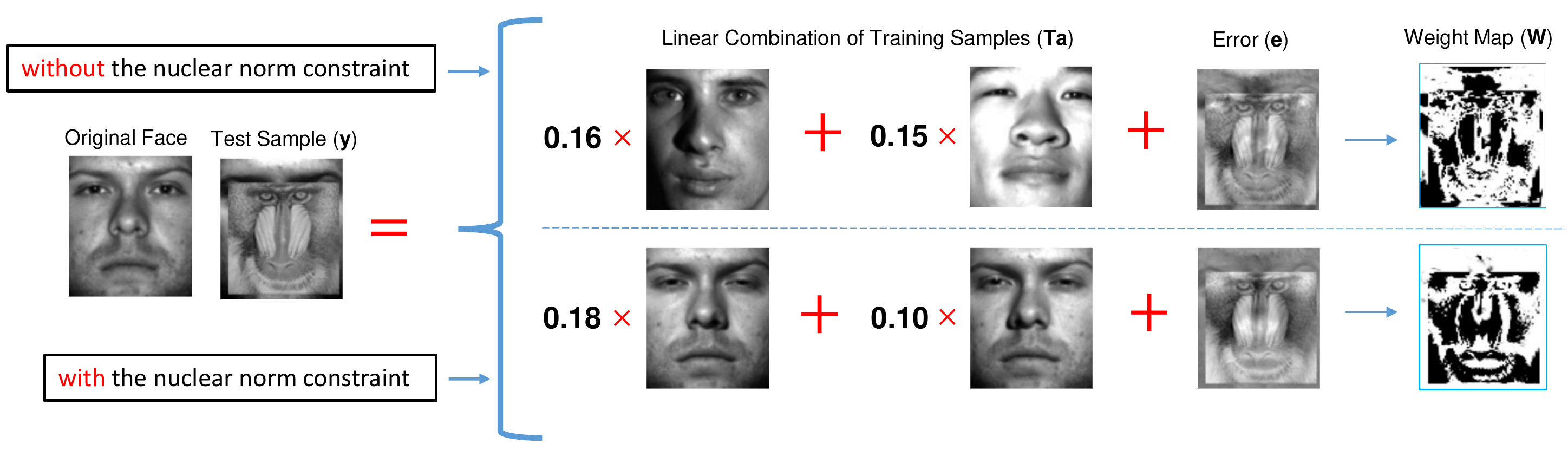}\\
     %\vspace{-0.3cm}
    \caption{An example of 70\% occluded face. If the low-rank constraint is not used (thus, allowing only the weighting norm in \eqref{eq:aug} to be utilized) then ${\bf y}$ is reconstructed from faces that belong to the wrong identities. On the other hand, our method which employs both the nuclear and weighting norms to describe the error, is able to reconstruct the occluded face from images that belong to the correct identity.}
     \vspace{-0.3cm}
     \label{fig:weight_explanation}
\end{figure*}

Notice that we have,
\begin{align}
J(\ba^t)&=J(\ba^t,\bw^{t+1})\nonumber\\
&\ge J(\ba^{t+1},\bw^{t+1})\ge J(\ba^{t+1},\bw^{t+2}) = J(\ba^{t+1}).
\end{align}

\subsection{Optimization}\label{sec:optim}

Let us now describe the iterative algorithm to find efficiently ${\bf {a}}^{t+1}$ in problem (\ref{eq:coding_lr_step}).

In order to solve the proposed problem, first we let ${\bf y - Ta = e}$ and since we are interested in estimating nonnegative coefficients for the representation vector we also introduce an additional variable ${\bf z}$ such that ${\bf a = z}$. Then, the coding step \eqref{eq:coding_lr_step} is reformulated as,
\begin{equation}
\begin{aligned}
\label{eq:constrained_lr_problem}
&\underset{{\bf a},{\bf z},{\bf e}}{\text{minimize}} \;\;\;  \Arrowvert \sqrt{{\bf W}^{t+1}}{\bf e} \Arrowvert_2^2 + \lambda_*\Arrowvert T_M({\bf e})\Arrowvert_* + \vartheta({\bf z}) \\
& \text{subject to \;\;}   {\bf y} - {\bf T}{\bf a} = {\bf e}, {\bf a = z}.
\end{aligned}
\end{equation}

Problem \eqref{eq:constrained_lr_problem} is solved efficiently with ADMM which is known for fast convergence to an approximate solution \cite{Boyd2011}. As in the method of multipliers, the problem takes the form of the augmented Lagrangian,
\begin{align}
\label{eq:lagrangian_lr}
& L({\bf e,a,z},{\bf u}_1,{\bf u}_2,{\bf w}^{t+1}) = \Arrowvert \sqrt{{\bf W}^{t+1}}{\bf e} \Arrowvert_2^2 + \lambda_*\Arrowvert T_M({\bf e})\Arrowvert_* \nonumber \\
&+ \vartheta({\bf z}) + {\bf u}_1^T\big({\bf y - Ta - e}\big) + \frac{\rho_1}{2}{\left\lVert {\bf y} - {\bf T}{\bf a} - {\bf e} \right\rVert}^2_2 \nonumber \\
& + {\bf u}_2^T\big({\bf a - z}\big) + \frac{\rho_2}{2}{\left\lVert {\bf a} - {\bf z} \right\rVert}^2_2,
\end{align}
where $\rho_1 > 0$ and $\rho_2 > 0$ are the penalty parameters, and ${\bf u}_1$ and ${\bf u}_2$ are the dual variables. The ADMM updates can be expressed as,
\begin{subequations}
    \begin{align}    
{ {\bf e}_{s+1}} &=  \argmin_{\bf e} L({\bf e},{\bf a}_{s},{\bf u}_{1,{s}},{\bf w}^{t+1}),  \\
{ {\bf z}_{s+1}} &=  \argmin_{\bf z} L({\bf z},{\bf a}_{s},{\bf u}_{2,{s}}),  \\
{ {\bf a}_{s+1}} &=  \argmin_{\bf a} L({\bf e}_{s+1},{\bf a},{\bf u}_{1,{s}},{\bf z}_{s+1},{\bf u}_{2,{s}}), \\
\label{eq:u1}
{\bf u}_{1,{s+1}} &=  {\bf u}_{1,s} + \rho_1({\bf y} - {\bf T}{\bf a}_{s+1} - {\bf e}_{s+1}),\\
\label{eq:u2}
{\bf u}_{2,{s+1}} &=  {\bf u}_{2,s} + \rho_2({\bf a}_{s+1} - {\bf z}_{s+1}),
 \end{align}
\label{eq:admmupdates}
\end{subequations}
where $s$ denotes the ADMM iteration and finally we set ${\bf a}^{t+1}=\lim_{s \to +\infty} {\bf a}_{s+1}$.

\subsubsection{Finding ${\bf e}_{s+1}$}\label{sec:e_update} The update of ${\bf e}_{s+1}$ is given by minimizing the following problem,
\begin{flalign}
{\bf e}_{s+1} &=  \argmin_{\bf e} \frac{1}{2}{\left\lVert {\bf y} - {\bf T}{\bf a}_{s} - {\bf e} \right\rVert}^2_2 + \frac{1}{\rho_1}\Arrowvert \sqrt{{\bf W}^{t+1}}{\bf e} \Arrowvert_2^2 \nonumber \\
&+ \frac{1}{\rho_1}{\bf u}_{1,s}^T\big({\bf y - T}{\bf a}_s - {\bf e}\big) + \frac{\lambda_*}{\rho_1}\Arrowvert T_M({\bf e}) \Arrowvert_*.
\label{eq:e_lr}
\end{flalign}
To calculate ${\bf e}_{s+1}$ we consider a two-step fast approximation. In step one we solve the weighted norm problem,
\begin{flalign}
{\tilde {\bf e}} &=  \argmin_{\bf e} \frac{1}{	2}{\left\lVert {\bf y} - {\bf T}{\bf a}_{s} - {\bf e} \right\rVert}^2_2 + \frac{1}{\rho_1}\Arrowvert \sqrt{{\bf W}^{t+1}}{\bf e} \Arrowvert_2^2 \nonumber \\
&+ \frac{1}{\rho_1}{\bf u}_{1,s}^T\big({\bf y - T}{\bf a}_s - {\bf e}\big),
\label{eq:e_lr_first}
\end{flalign}
and in step two to satisfy the nuclear norm constraint we project the estimated ${\tilde {\bf e}}$ to a low-rank space, to obtain,
\begin{equation}
{{\bf e}_{s+1}} = \argmin_{\bf e} \frac{1}{2}{\left\lVert T_M\Big({\bf e} - {\tilde {\bf e}}\Big) \right\rVert}^2_F + \frac{\lambda_*}{\rho_1}\Arrowvert T_M({\bf e}) \Arrowvert_*.
\label{eq:e_lr_second}
\end{equation}
Problem \eqref{eq:e_lr_first} has a closed-form solution given by,
\begin{flalign}
{\tilde {\bf e}} = {\bf C}^{-1}\big({\bf y} - {\bf T}{\bf a}_{s} + {\bf u}_{1,s}/\rho_1\big),
\label{eq:e_proximity}
\end{flalign}
where ${\bf C} = \big( {\bf I} + 2{\bf W}^{t+1}/\rho_1\big)$ is a diagonal matrix with diagonal entries $c_i = 1 + 2w_{i}^{t+1}/\rho_1$. Since {\bf C} is a diagonal matrix, to update ${\tilde {\bf e}}$ we only need to construct a vector with elements equal to $1/c_i$ and perform an element wise-multiplication between the constructed vector and the residual vector ${\bf y} - {\bf T}{\bf a}_{s} + {\bf u}_{1,s}/\rho_1$. Thus, the update of ${\tilde {\bf e}}$ can be calculated fast. 

The update in \eqref{eq:e_proximity} is in essence an outlier detector similar to the {\em{soft-thresholding}} operator \cite{Boyd2011}. The values of the residual vector ${\bf y} - {\bf T}{\bf a}_{s} + {\bf u}_{1,s}/\rho_1$ will be weighted according to ${\bf C}^{-1}$. A small weight (close to zero) will be given to non-outlier elements (e.g., elements of the residual vector with small values) while a large weight (close to one) will be given to outliers (e.g., elements of the residual vector with large values).   

Problem \eqref{eq:e_lr_second} is a nuclear norm minimization problem of the form,  
\begin{equation}
\min_{\bf X} \frac{1}{2}{\left\lVert {\bf X} - \Delta \right\rVert}^2_F +  \lambda_*{\Arrowvert{{\bf X}}\Arrowvert_*}, 
\label{eq:proximity_lr}
\end{equation}
which has a closed-form solution given by \cite{Boyd2011},
\begin{equation}
{\hat {\bf X}} = \mathcal{L}_{\lambda_*}(\Delta) = {\bf U}_\Delta\mathcal{S}_{\lambda_*}({\bf \Sigma}_{\Delta}){\bf V}_\Delta^T,\\
\end{equation}
where $\mathcal{S}_{\lambda_*}({\bf \Sigma}_{\Delta}) = \text{sign}(\delta_{ij})\text{max}(0,|\delta_{ij}| - \lambda_*)$ is the soft-thresholding operator, $\mathcal{L}_{\lambda_*}(\cdot)$ is the singular value soft-thresholding operator and $\Delta = {\bf U}_\Delta{\bf \Sigma}_{\Delta}{\bf V}_\Delta^T$ is the SVD of matrix $\Delta$. 

Thus, the two-step solution of \eqref{eq:e_lr} is given by, 
\begin{subequations}
\label{eq:nn_proximity}
\begin{flalign}
\label{eq:e_lr_first_step}
&{\tilde {\bf e}} = {\bf C}^{-1}\big({\bf y} - {\bf T}{\bf a}_{s} + {\bf u}_{1,s}/\rho_1\big)\\
\label{eq:e_lr_second_step}
&{{\bf e}_{s+1}} = \mathcal{L}_{\lambda_* / \rho_1}\big(T_M(\tilde{\bf e})\big).
\end{flalign}
\end{subequations}
The solution in \eqref{eq:nn_proximity} is the low-rank estimation of the weighted error image \footnote{Note that the solution in \eqref{eq:e_lr_second_step} is obtained in a matrix form which is then transformed to a column-wise vectorized form ${{\bf e}_{s+1}}$.}.

\subsubsection{Finding ${\bf z}_{s+1}$}\label{sec:z_update} The update of ${\bf z}_{s+1}$ is obtained by solving the following problem,
\begin{align}
{{\bf z}_{s+1}} = \argmin_{\bf z}\frac{1}{2}{\left\lVert {\bf a}_s - {\bf z} \right\rVert}^2_2 + \frac{1}{\rho_2}\vartheta({\bf z}) + \frac{1}{\rho_2}{\bf u}_{2,s}^T\big({\bf a}_s - {\bf z}\big).
\label{eq:z}
\end{align}
The solution of \eqref{eq:z} is given by,
\begin{flalign}
{{\bf z}_{s+1}} = ({\bf a}_s + {\bf u}_{2,s}/\rho_2)_+,
\label{eq:z_proximity}
\end{flalign}
where $(\cdot)_+$ is the function the keeps only the positive coefficients of its argument and set the rest to zero. 

Notice that we only need to change the update of ${\bf z}$ for different regularization functions $\vartheta({\bf z})$. For example, to solve an iterative reweighted sparse coding (IRSC) problem and regularize the coefficients to be sparse ($\vartheta({\bf z}) = \lambda||{\bf z}||_1$) we have to substitute \eqref{eq:z_proximity} with the {\em{soft-thresholding}} operator \cite{Boyd2011}. To solve an iterative reweighted least squares (IRLS) problem ($\vartheta({\bf a}) = \lambda||{\bf a}||^2_2$) we do not need to introduce and estimate the additional variables, ${\bf z}$ and ${\bf u}_2$. The coefficients ${\bf a}$ can be estimated by solving a regularized least squares problem.   

\subsubsection{Finding ${{\bf a}_{s+1}}$}\label{sec:a_update} The update of ${\bf a}_{s+1}$ is obtained by solving the following problem, 
\begin{flalign}
{{\bf a}_{s+1}} &=  \argmin_{\bf a} \frac{1}{2}{\left\lVert {\bf y} - {\bf Ta} - {\bf e}_{s+1}\right\rVert}^2_2 + \frac{1}{\rho_1}{\bf u}_{1,s}^T\big({\bf y - T}{\bf a} - {\bf e}_{s+1}\big) \nonumber \\
& + \frac{1}{\rho_1}{\bf u}_{2,s}^T\big({\bf a} - {\bf z}_{s+1}\big) + \frac{\rho_2}{2\rho_1}{\left\lVert {\bf a} - {\bf z}_{s+1}\right\rVert}^2_2.
\label{eq:a}
\end{flalign}
Notice that equation \eqref{eq:a} is a regularized least squares problem, whose closed-form solution is, 
\begin{equation}
\label{eq:a_closed}
{{\bf a}_{s+1}} = {\bf P}\Big({\bf T}^T({\bf y} - {\bf e}_{s+1} + {\bf u}_{1,s}/\rho_1) + (\rho_2/\rho_1){\bf z} - {\bf u}_{2,s}/\rho_1\Big),
\end{equation}
where ${\bf P} = \left({\bf T}^T {\bf T} + \frac{\rho_2}{\rho_1}{\bf I}\right) ^{-1}$ and can be pre-calculated once and cached offline. For large matrices ${\bf P}$, iterative algorithms can be used for solving  this linear system of equations when matrix inversion is not feasible.

An example of our approach is presented in Figure~\ref{fig:weight_explanation}. If the low-rank constraint is not used then ${\bf y}$ is reconstructed using faces that belong to the wrong identities. On the other hand, our approach with the use of the nuclear norm constraint is able to reconstruct the occluded face using images that belong to the correct identity (second row in Figure~\ref{fig:weight_explanation}). Also, notice the differences in weight map estimations. In the first case small weights (black values) assigned without any structure to any region of the face. This is not desirable since informative pixels were detected as outliers (e.g., pixels around the occluded object). In addition, many pixels on the occluded object were detected as inliers. In the second case, the error is low-rank and has a spatial continuity around the occluded object (pixels close to zero). In this case, the weight map ${\bf W}$ is also enforced to have a spatial continuity (since ${\bf W}$ is related to the error) with small weights assigned to the occluded object as desired.

We would like to mention here a closely related work for block occlusion errors namely robust low-rank regularized regression ($\text{RLR}^3$)\cite{Qian2014}. There are two fundamental differences between our work and $\text{RLR}^3$. First, the weighted residual ${\bf W}({\bf y - Ta})$ instead of the residual error ${\bf y - Ta}$ was modeled to be low-rank which is different from our method. The model in $\text{RLR}^3$ can handle occlusion of specific objects and size that covers a portion of the face image entirely from left to right (or from top to bottom) such as scarves. Our method handles block occlusions that appear in any size and place in the face. Second, the function to be minimized in $\text{RLR}^3$ is not derived from the duality theorem \cite{Rockafellar:1996} which raises concerns about its convergence guarantees. 

The complete steps of the fast low-rank and iterative reweighted nonnegative least squares (F-LR-IRNNLS) algorithm for contiguous errors are presented in Algorithm~\ref{alg:firc}. 

% Algorithm 2: The FIRC Proposed Algorithm
\begin{algorithm}[!t]
\caption{Fast \& Low-Rank IRNNLS Algorithm}
\label{alg:firc}
\begin{algorithmic}[0]
\State {\textbf{Inputs}}: ${\bf y}$, ${\bf T}$, $\lambda_*$,  $\rho_1$, $\rho_2$, $\epsilon_1$, $\epsilon_2$ and $\epsilon_3$.
%\State {Calculate ${\bf y_c} = LRR({\bf y,T})$}
\State {Initialize ${\bf a}^1 = 1/n$, ${\bf u}_{1,1} = {\bf 0}$, ${\bf u}_{2,1} = {\bf 0}$, $t=0$ and $s=0$} 
\State {\textbf{Repeat}}
\begin{enumerate}[leftmargin=0.5cm]
\item $t = t + 1$ 
\item Estimate the weights,
\begin{equation}
{w_i^t} = {\phi'(({\by- \bT\ba^t})_i)/({\by- \bT\ba^t})_i}, \;\; i=1,\ldots,d \nonumber
\end{equation}

\end{enumerate}

\State\hspace{\algorithmicindent} {\textbf{Repeat} }

\begin{enumerate}[leftmargin=1.5cm]
\setcounter{enumi}{2}
\item $s = s + 1$
\item Find ${{\bf e}_s}$ using \eqref{eq:nn_proximity} (contiguous errors) or \eqref{eq:e_fast_proximity} (non-contiguous errors)
\item Find ${{\bf z}_s}$ using \eqref{eq:z_proximity} 
\item Update ${{\bf a}_s}$ using \eqref{eq:a_closed} 
\item Update ${ {\bf u}_{1,s}}$ and ${ {\bf u}_{2,s}}$ using \eqref{eq:u1}, \eqref{eq:u2} 
\end{enumerate}

\State\hspace{\algorithmicindent} {\textbf{Until} {\em converge}} %${\left\lVert {\bf y} - {\bf Ta}^{s+1} - {\bf e}^{s+1} \right\rVert}_\infty \leq \epsilon_1$ %\& ${\left\lVert {\bf a}^{s+1} - {\bf z}^{s+1} \right\rVert}_\infty \leq \epsilon_3$

\begin{enumerate}[leftmargin=0.5cm]
\setcounter{enumi}{7}
\item Set ${\bf a}^{t} = {\bf a}_{s}$, ${\bf u}_{1,1} = {\bf 0}$, ${\bf u}_{2,1} = {\bf 0}$, $s=0$
\end{enumerate} 

\State {\textbf{Until}} {\em converge}

\State {\textbf{Output}}: The final estimates of ${\bf a}$ and ${\bf w}$.

\end{algorithmic}
\end{algorithm}

\subsection{Identification Scheme}\label{sec:classification}

In SRC \cite{Wright2009},  the face class that yields the minimum reconstruction error is selected in order to classify or identify the subject. Similarly, in this work the classification is given by computing the residuals $e$ for each class $i$ as,

\begin{equation}
e_{i}({\bf y})={\Arrowvert \sqrt{{\bf W}^{f}} ({\bf y} - T_{i}{\hat{\bf a}}_{i})\Arrowvert _{2}},
\label{eq:rechg}
\end{equation}
where ${\hat{\bf a}}_{i}$ is the segment of the final estimated ${\bf a}$ associated with class $i$ and ${\bf W}^f$ is the final estimated weight matrix from Algorithm~\ref{alg:firc}. Finally, the identity of ${\bf y}$ is given as, $\text{Identity}({\bf y})=\argmin_{i}\{ e_{i}({\bf y})\}$.

\subsection{The Weight Function}\label{sec:weights}

Ideally, the weight function should distinguish inliers and outliers given a training dictionary with non-occluded faces and a test sample with occlusion \cite{Yang2013}. In particular, given the residual error at any iteration, small weights (close to zero) should be assigned to the outlier pixels (large residual error) and larger weight (close to one) to the inlier pixels (small residual error). Although any weight function \cite{Huber1981,He2014} of the form $w = \phi'(x)/x$ can be used in our framework as long as $\phi'(x)/x$ is decreasing on $(0,\infty)$, in this work we utilize the logistic function proposed in \cite{Yang2013}. The logistic function performs particularly well in FI\footnote{For a complete justification of the effectiveness of the logistic function in FI we refer the reader to \cite{Yang2013}.}. The weight component $w_i$ as a function of $x_i$, which is decreasing on $(0,\infty)$, is given by,
\begin{equation} 
      {{w_i}} = \frac{\exp \Big( -\mu\Big(x_i\Big)^2 + \mu\eta\Big)}{1+\exp \Big( -\mu\Big(x_i\Big)^2 + \mu\eta\Big)}, \;\; i=1,\ldots,d,
\label{eq:logistic}
\end{equation}
where $\mu$ and $\eta$ are positive scalars.  As in \cite{Yang2013}, $\eta$ denotes the value of the $l^{th}$ largest element of the residual vector ${\bf x}$, where $l = \floor*{\gamma d}$, $\gamma \in (0,1)$. $\mu$ is given as $\frac{\zeta}{\eta}$ with $\zeta = 8$. We also set $\gamma = 0.8$ for the experiments without occlusion and $\gamma = 0.6$ for the experiments with occlusion as in \cite{Yang2013}.

\begin{figure}[!t]
\centering          
\begin{subfigure}[b]{0.25\textwidth}\includegraphics[scale=0.29]{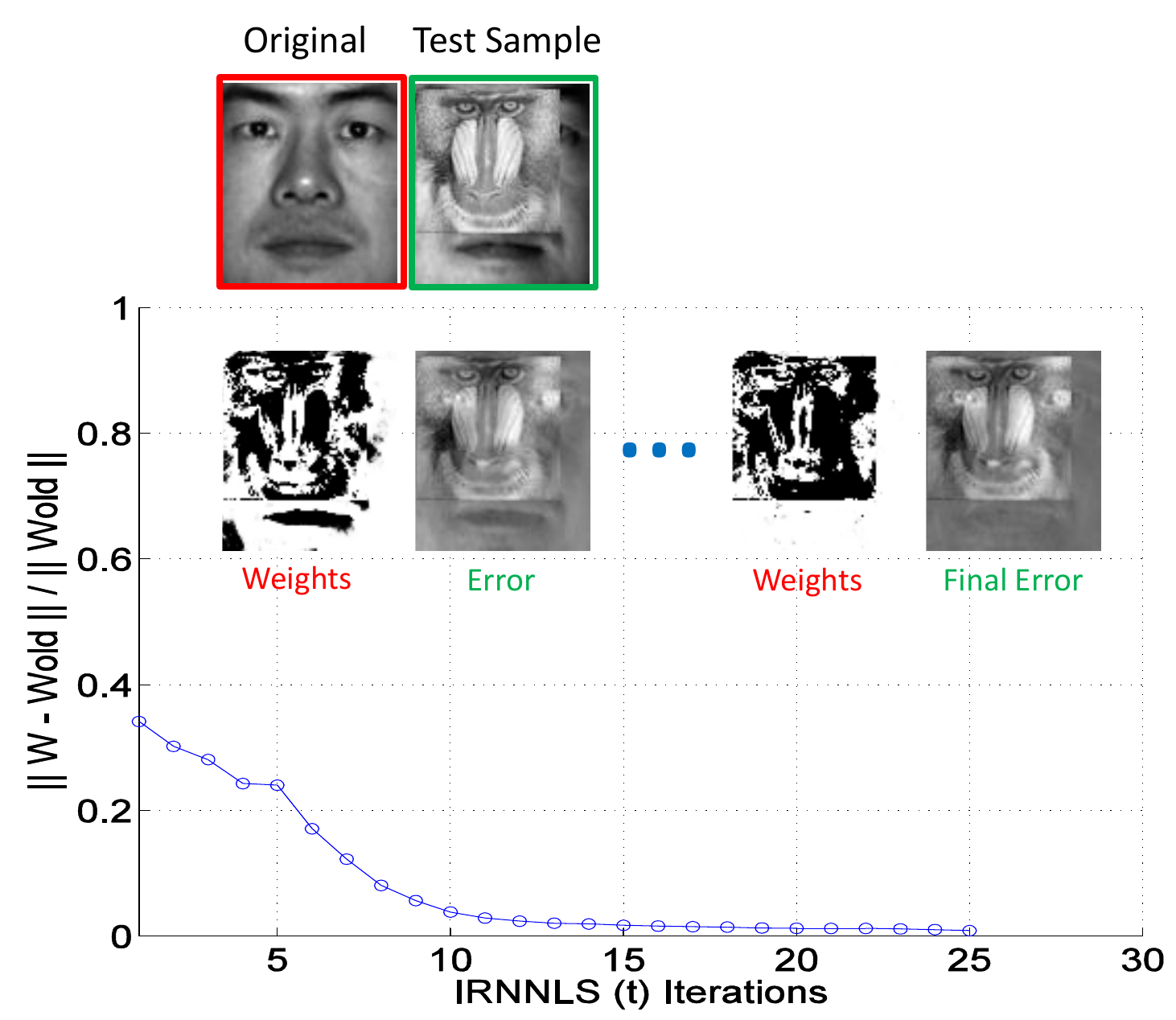} \label{fig:convergence_left_results}
\end{subfigure}%\hspace{0.1cm}
\begin{subfigure}[b]{0.25\textwidth}\includegraphics[scale=0.24]{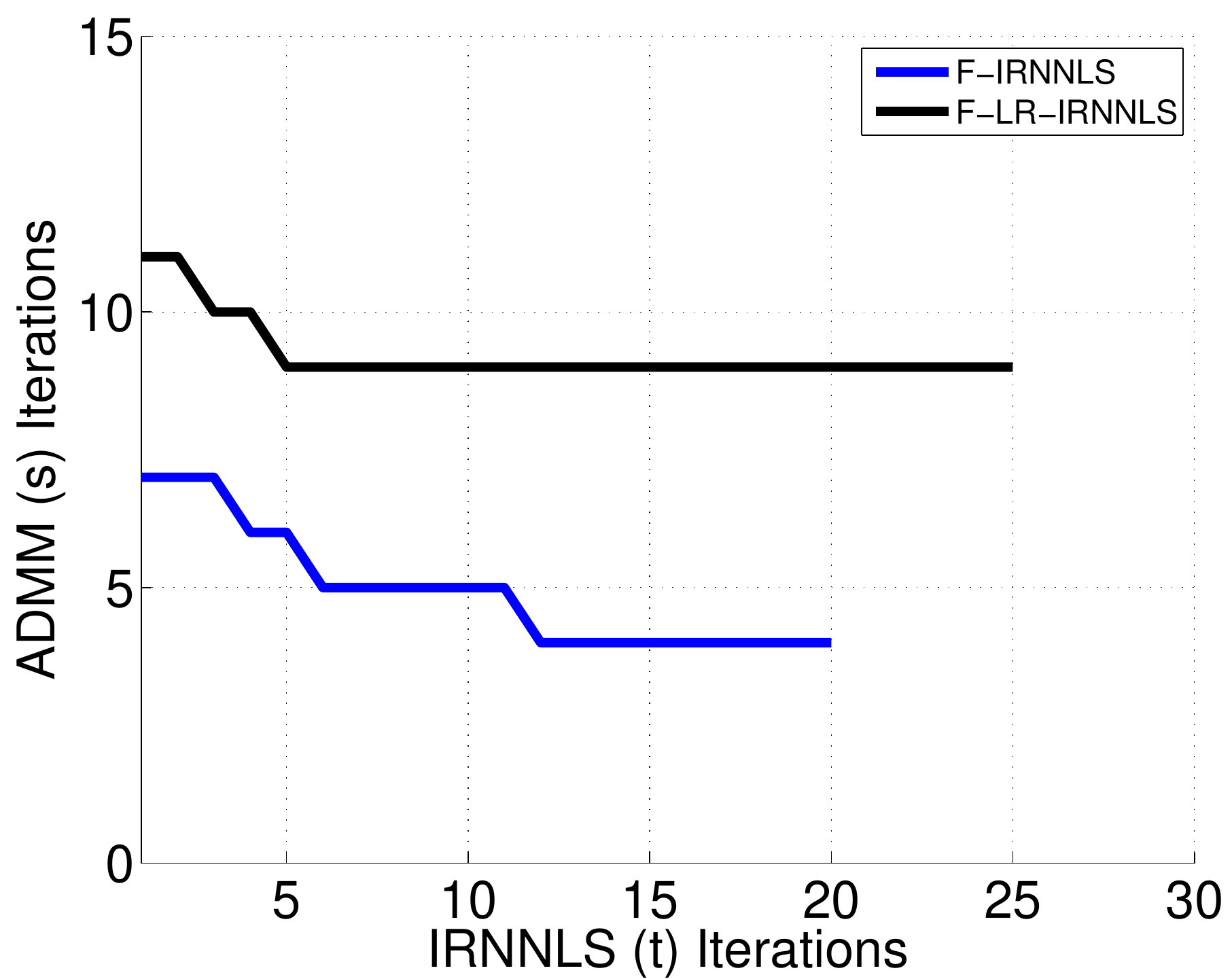}\label{fig:convergence_right_results}
\end{subfigure}%\hspace{0.1cm}
\caption{The left graph shows the convergence of the iterative reweighted and low-rank algorithm (F-LR-IRNNLS). In particular, we present the change of the weights between two consecutive iterations. As $t\to\infty$ the error becomes more sparse and structural, thus, small weights are concentrated on the occluding object. The right graph shows the number of ADMM iterations as a function of the reweighted iterations.} 
\label{fig:method_convergence}
\vspace{-0.3cm}
\end{figure}

\subsection{Convergence Criteria}\label{sec:convergence}

For the purpose of this paper, in order to guarantee convergence of the optimization problem \eqref{eq:constrained_lr_problem} using ADMM, it is sufficient to enforce appropriate termination criteria. As suggested in \cite{Boyd2011}, we enforced the primal residuals to be small such that ${\left\lVert {\bf y} - {\bf Ta} - {\bf e} \right\rVert}_2 \leq \epsilon_1$ and ${\left\lVert {\bf a} - {\bf z} \right\rVert}_2 \leq \epsilon_2$, where $\epsilon_1$ and $\epsilon_2$ are small positive numbers.

The termination criterion for the iterative reweighted sequences is ${\left\Vert {\bf W}^{t} - {\bf W}^{t-1} \right\Vert _{2}}/{\left\Vert {\bf W}^{t-1} \right\Vert _{2}}< \epsilon_3$, where $\epsilon_3$ is a small positive number. Figure~\ref{fig:method_convergence} (left) shows the change of the weights between two consecutive iterations for one particular example. 

\subsection{Special case: robust representation for non-contiguous errors}

A special case of our method leads to the robust representation method \cite{Yang2011,Yang2013,He2014,He2011} when used for FI problem with non-contiguous errors as given in \eqref{eq:ja}. In the previous robust methods \cite{Yang2011,Yang2013,He2014,He2011} high computational cost is exhibited due to the reweighted coding step to estimate ${\bf a}$. To address this issue the proposed ADMM algorithm described above is utilized here as well to solve this case efficiently. The augmented function of the robust representation problem is formulated as,
\begin{equation}
    \begin{aligned}    
\label{eq:fastaug}
{J}({\bf {a}},{\bf {w}}) = \frac{1}{2}\Arrowvert \sqrt{\bf W}({\bf y}-{\bf Ta})\Arrowvert _{2}^{2} + {\varphi({\bf w})} + \vartheta{\left({\bf a}\right)},
 \end{aligned}
\end{equation}
which is similar to \eqref{eq:aug} but without the nuclear norm term. Similarly to \eqref{eq:aug}, a local minimizer $({{\bf a},{\bf w}})$  can be calculated in two steps,
 \begin{subequations}
\label{eq:irls_first}  
   \begin{align}    
\label{eq:weight_step}
{w_i^{t+1}} &= \phi'(({\by- \bT\ba^t})_i)/({\by- \bT\ba^t})_i\\
\label{eq:coding_step}
{ {\bf \hat a}}^{t+1} &= \argmin_{\bf a}\Arrowvert \sqrt{{\bf W}^{t+1}}({\bf y}-{\bf Ta})\Arrowvert _{2}^{2}  + \vartheta{\left({\bf a}\right)}.
\end{align} 
\end{subequations}
A major drawback of the iterative reweighted algorithm is that it is computationally expensive \cite{Zhou2014} due to the coding step \eqref{eq:coding_step}. High computational cost is exhibited regardless of the coefficient regularization, for example $\vartheta{\left({\bf a}\right)} = \lambda||{\bf a}||_2^2$, $\vartheta{\left({\bf a}\right)} = \lambda||{\bf a}||_1$ or $\vartheta{\left({\bf a}\right)}$ is the indicator function of the nonnegative orthant $\mathbb{R}^n_+$ \cite{Yang2011,Yang2013,He2014,He2011}. The coding step is expensive since in each iteration a new weighted system matrix $\sqrt{{\bf W}^{t+1}}{\bf T}$ is provided given the updated weights. Moreover, an offline pre-calculation of the weighted inverse is not possible. Efficient methods to solve problems in the form of \eqref{eq:coding_step} such as conjugate gradient \cite{Shewchuk1994} or $\ell_1$ algorithms \cite{Yang2010,Babacan2010} may require several iterations to converge to the desired point \cite{Yang2013}. The active set method \cite{Lawson1995} used for solving the nonnegative least squares problem becomes also slow due to the computation of the pseudoinverse of the system matrix in each iteration. 

%The idea is based on \cite{Zhou2014} where the iterative %reweighted scheme with weights applied to the regularization %term $\vartheta{\left({\bf a}\right)}$ rather to the residual %was accelerated by reformulating the problem and using ADMM to %optimize it

To solve \eqref{eq:coding_step} efficiently for the $\vartheta{\left({\bf a}\right)}$ functions described earlier we utilize the proposed ADMM algorithm described above. Thus, we avoid the explicit calculation of ${\bf T}^T{\bf W}^{t+1}{\bf T}$ inverse during the execution of the algorithm. The idea of accelerating the iterative reweighed scheme using ADMM is also found in \cite{Zhou2014}. However, in \cite{Zhou2014} the weights applied to the regularization term $\vartheta{\left({\bf a}\right)}$ rather to the residual as in done here. To accelerate \eqref{eq:coding_step}, we set ${\bf y - Ta = e}$, ${\bf a = z}$ and the problem is reformulated as,
\begin{equation}
\begin{aligned}
\label{eq:constrained_problem}
&\underset{{\bf a},{\bf z},{\bf e}}{\text{minimize}} \;\;\;  \Arrowvert \sqrt{{\bf W}^{t+1}}{\bf e} \Arrowvert_2^2  + \vartheta({\bf z}) \\
& \text{subject to \;\;}   {\bf y} - {\bf T}{\bf a} = {\bf e}, {\bf a = z}.
\end{aligned}
\end{equation}
Notice that the $\sqrt{{\bf W}^{t+1}}{\bf T}$ term is no longer part of the optimization problem which allows us to solve \eqref{eq:constrained_problem} efficiently. 

Problem \eqref{eq:constrained_problem} has similar ADMM updates with \eqref{eq:constrained_lr_problem} except ${\bf e}_{s+1}$. To find ${\bf e}_{s+1}$ we only have to calculate, 
\begin{flalign}
{{\bf e}_{s+1}} = {\bf C}^{-1}\big({\bf y} - {\bf T}{\bf a}_s + {\bf u}_{1,s}/\rho_1\big),
\label{eq:e_fast_proximity}
\end{flalign}
where ${\bf C}$ is the diagonal matrix defined in \eqref{eq:e_proximity} and $ {\bf C}^{-1}$ can be calculated fast due to the diagonal structure. The updates of ${\bf z}_{s+1}$ and ${\bf a}_{s+1}$ are similar to \eqref{eq:z_proximity} and \eqref{eq:a_closed} respectively and as explained earlier ${\bf P}$ matrix can be pre-calculated once and cached offline\footnote{As noted earlier, the offline calculation of ${\bf P}$ can be utilized when direct inversion is feasible.}. Thus, to solve the robust representation problem we utilize an efficient method since no online inversion of the system matrix is performed in any of the variable updates.

Our fast iterative reweighted nonnegative least squares (F-IRNNLS) algorithm just described solves the robust representation problem efficiently and the steps are also presented in Algorithm~\ref{alg:firc}.

The number of ADMM iterations required for each reweighted iteration is presented in Figure~\ref{fig:method_convergence} (right) for our method. As expected the number of ADMM iterations for F-LR-IRNNLS is greater than the F-IRNNLS due to the calculation of the nuclear term.

\subsection{Dealing with corrupted testing and training samples}\label{sec:corrupted_training}

Until now we have assumed that the training samples represent ``clean" frontal aligned views and without large variations of the same identity. However, there might be scenarios such as face identification in an unconstrained environment where {\em testing} as well as {\em training} samples are occluded. 

To deal with this scenario we incorporate into our framework techniques based on RPCA \cite{Cand`es2011} to separate outlier pixels and occlusions from the training samples. The main idea of RPCA is that the training matrix is decomposed into ${\bf T = A + E}$, where ${\bf A}$ denotes the dictionary with clean faces and ${\bf E}$ is the remaining error matrix. Out of the many extensions of RPCA such as in \cite{Chen2012,Chen2014,Zhang2013,Liu2013,Zhang2015}, in this work we utilize the SLR method in \cite{Jiang2015} to separate outlier pixels since it is robust to variations such as expression and pose which are very common in an unconstrained environment. One of the main differences between the RPCA \cite{Cand`es2011} and SLR \cite{Jiang2015} methods is that an additional intra-class variation dictionary ${\bf B}$ is estimated in the latter method using ${\bf T = A + B + E}$. Thus, having estimated dictionaries ${\bf A}$ and ${\bf B}$ by \cite{Jiang2015} we can modify our degradation model \eqref{eq:deg_model} to,
\begin{align}
{\bf y} = {\bf A}{\bf a}_1 + {\bf B}{\bf a}_2 + {\bf e},
\label{eq:unconstrained_model}
\end{align}
where ${\bf A} \in \mathbb{R}^{d \times n}$ denotes a class-specific dictionary (clean images dictionary) and ${\bf B} \in \mathbb{R}^{d \times n}$ a non-class specific dictionary (intra-class variation dictionary). Also, ${\bf a}_1 \in \mathbb{R}^n$ and ${\bf a}_2 \in \mathbb{R}^n$ are the representation vectors for the ${\bf A}$ and ${\bf B}$ dictionaries respectively and ${\bf e}$ is the representation error described earlier in our method. More compactly, the model in \eqref{eq:unconstrained_model} can be written as,
\begin{align}
{\bf y} = {\bf {T^\prime}{a^\prime}} + {\bf e},
\label{eq:unconstrained_model_compact}
\end{align}
where ${\bf {T^\prime}} = [ {\bf A \;\; B} ] \in \mathbb{R}^{d \times 2n}$ and ${\bf {a^\prime}} = [ {\bf a}_1 \; {\bf a}_2 ]  \in \mathbb{R}^{2n}$. Dictionary ${\bf B}$ in this case captures variations such as in expressions and pose that cannot be represented by the error term ${\bf e}$. On the other hand, ${\bf e}$ is utilized to capture occlusions and low-rank variations of the test image, as explained earlier in this work, that cannot be captured by ${\bf B}$. For the problem described above the function to be minimized is the same as in \eqref{eq:philr} where ${\bf T^\prime}$ and ${\bf a^\prime}$ are used instead of ${\bf T}$ and ${\bf a}$, respectively. Furthermore, the calculation of ${\bf A}$ and ${\bf B}$ is described in \cite{Jiang2015}. In our experiments we denote by F-LR-IRNNLS (SLR), the method when the SLR algorithm \cite{Jiang2015} is employed as a pre-processing step in our F-LR-IRNNLS method.

\section{Experimental Results}\label{sec:exps}

In this section we present experiments on four publicly available databases, AR \cite{Martinez1998}, Extended Yale B \cite{Georghiades2001}, Multi-PIE \cite{Gross2010} and Labeled Faces in the Wild (LFW) \cite{Huang2007} to show the efficacy of the proposed method. We demonstrate identification and reconstruction results under various artificial and real-world variations. We compare our framework with ten other FI algorithms, SRC \cite{Wright2009}, CR-RLS \cite{Zhang2011}, $\text{LR}^3$ \cite{Qian2014}, \L12 \footnote{This method solves the problem, $\min_{\bf a} \left\Vert T_M({\bf y}-{\bf Ta})\right\Vert _{12} + \lambda\left\Vert {\bf a}\right\Vert _{2}^{2}$. The employment of the $\ell_2$ norm for the coefficients was chosen to make fair comparisons with CR-RLS \cite{Zhang2011} and $\text{LR}^3$ \cite{Qian2014}. }, and the robust algorithms SDR-SLR \cite{Jiang2015}, HQ (additive form) \cite{He2014}, CESR \cite{He2011}, RRC\_L1 \cite{Yang2013}, RRC\_L2 \cite{Yang2013}, SSEC \cite{Li2013}. We consider the following five FI cases:
\begin{enumerate}
\item cases with contiguous variations such as random block occlusion with different sizes and objects, face disguise and mixture noise which is a combination of block occlusion and pixel corruption,
\item cases with non-contiguous variations such as illumination variations, pixel corruption, face expressions. 
\item cases with random block occlusion and few training samples. 
\item cases in unconstrained environment and with corrupted testing and training samples. 
\end{enumerate}
For all methods, we used the solvers provided by the authors of the corresponding papers. We chose to solve the $\ell_1$ minimization problem in SRC and RRC\_L1 with the Homotopy algorithm\footnote{The source code of Homotopy algorithm can be downloaded at http://www.eecs.berkeley.edu/~yang/software/l1benchmark/} \cite{Donoho2008} since it resulted in the highest accuracy in the performance comparison in \cite{Yang2010} with reasonable time execution. In our algorithms, we set $\lambda_* = 0.05$, $\rho_1 = 1$ and $\rho_2 = 0.1$. The convergence parameters were set equal to $\epsilon_1 = 10^{-2}$, $\epsilon_2 = 10^{-1}$, $\epsilon_3 = 10^{-2}$. For fair comparisons with respect to execution time and identification rates we set the same $\epsilon_3$ for the RRC algorithm and the same maximum number of iterations ($t = 100$). All face images were normalized to have unit $\ell_2$-norm and all variables initialized to zero except for ${\bf a}^1 = 1/n$ as in \cite{Yang2013}.

\subsection{Identification under Block Occlusions}

Experiments with occluded images were conducted on three datasets: Extended Yale B, AR and Multi-PIE.

\begin{figure}[!t]
\centering          
\begin{subfigure}[b]{0.10\textwidth}\includegraphics[width=\textwidth]{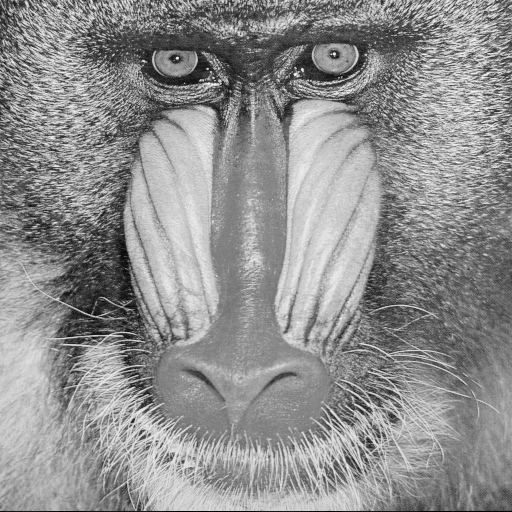}
\caption{Baboon}
\end{subfigure}%\hspace{0.1cm}
\begin{subfigure}[b]{0.18\textwidth}\includegraphics[width=\textwidth]{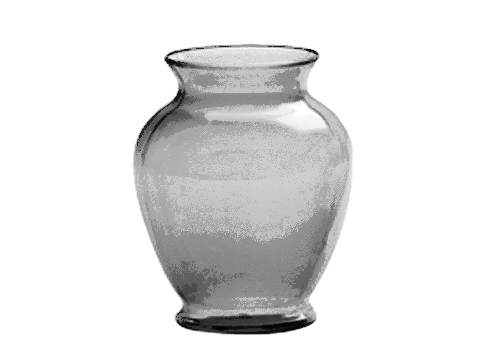}
\caption{Non-square image }\label{fig:occluded_vase}
\end{subfigure}%\hspace{0.1cm}
\begin{subfigure}[b]{0.10\textwidth}\includegraphics[width=\textwidth]{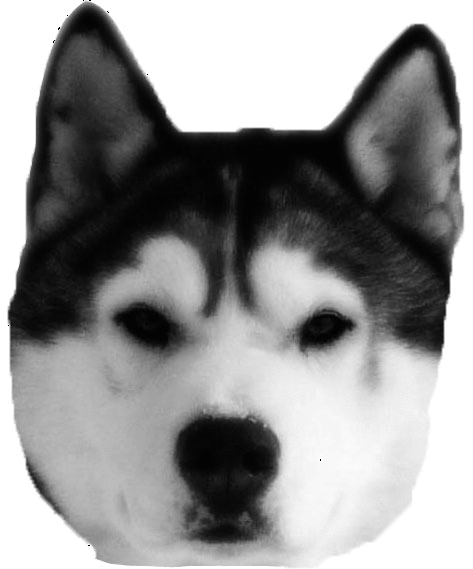}
\caption{Dog}\label{fig:occluded_dog}
\end{subfigure}%\hspace{0.1cm}
\caption{The three artificial images used for the block occlusion experiments.} 
\label{fig:occluded_images}
\vspace{-0.5cm}
\end{figure}

\begin{figure*}[!t]
\centering          
\begin{subfigure}[b]{0.25\textwidth}\includegraphics[width=\textwidth]{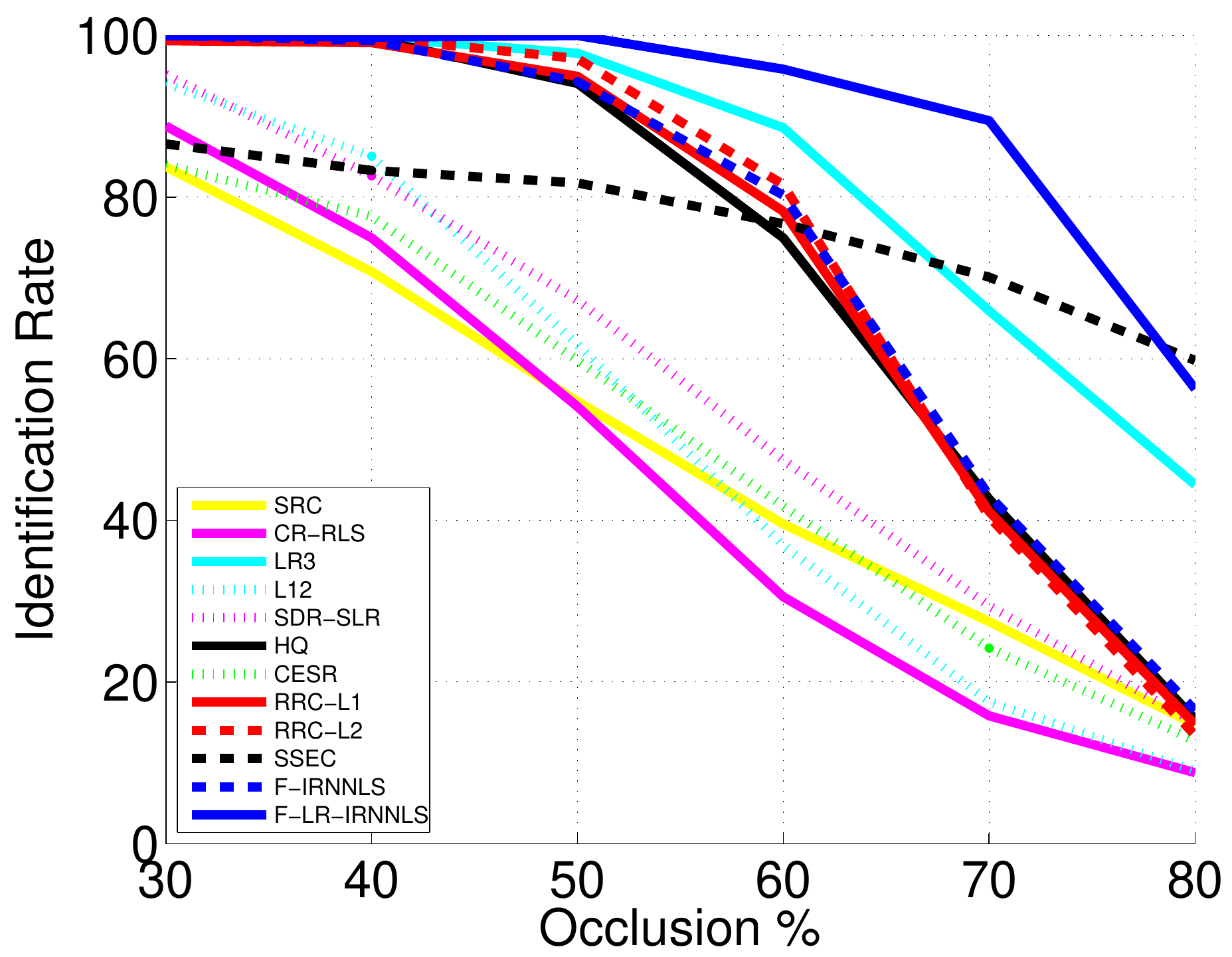}
\caption{Yale B dataset with baboon image}\label{fig:block_baboon}
\end{subfigure}%\hspace{0.1cm}
\begin{subfigure}[b]{0.25\textwidth}\includegraphics[width=\textwidth]{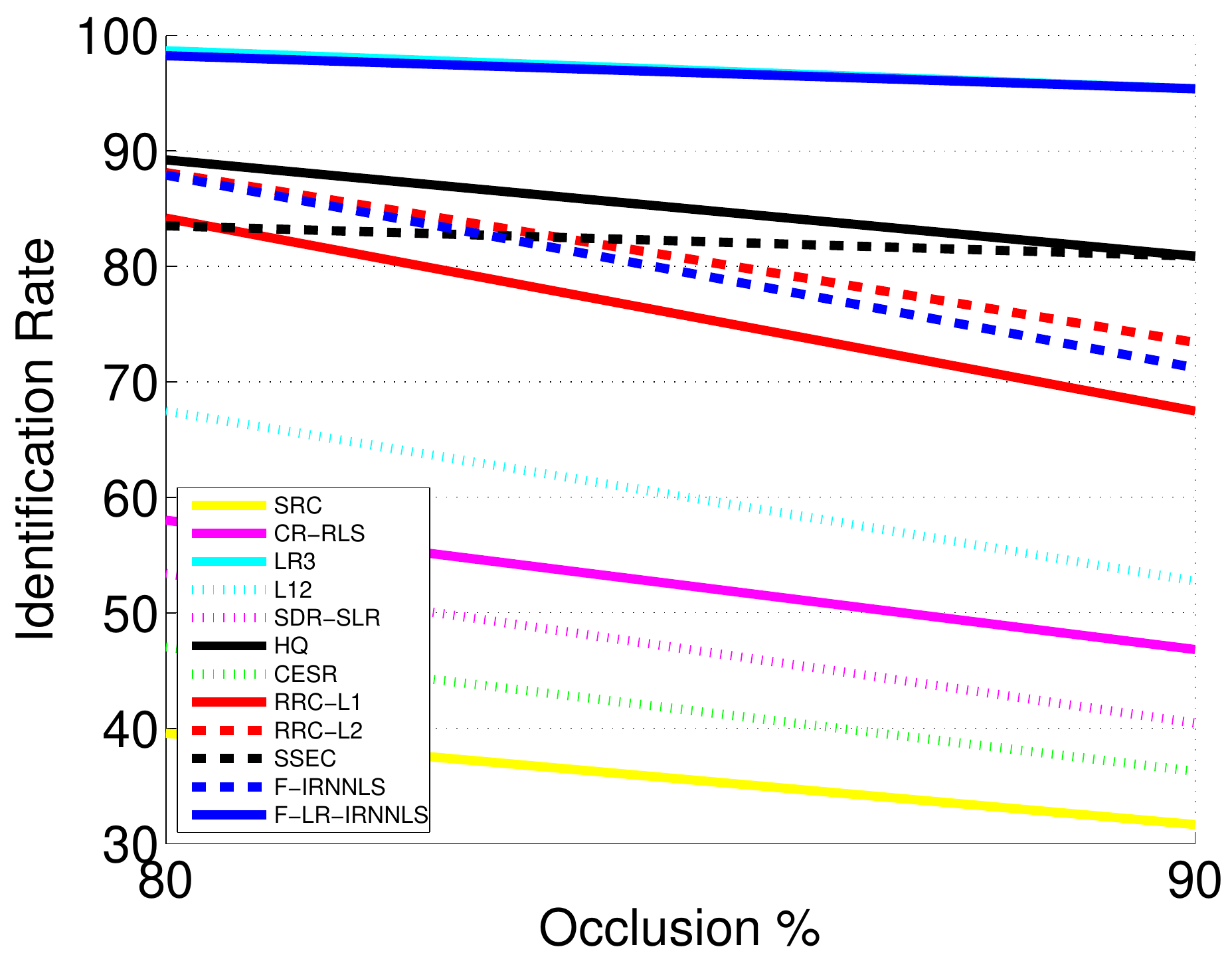}
\caption{Yale B dataset with vase image }\label{fig:block_vase}
\end{subfigure}%\hspace{0.1cm}
\begin{subfigure}[b]{0.25\textwidth}\includegraphics[width=\textwidth]{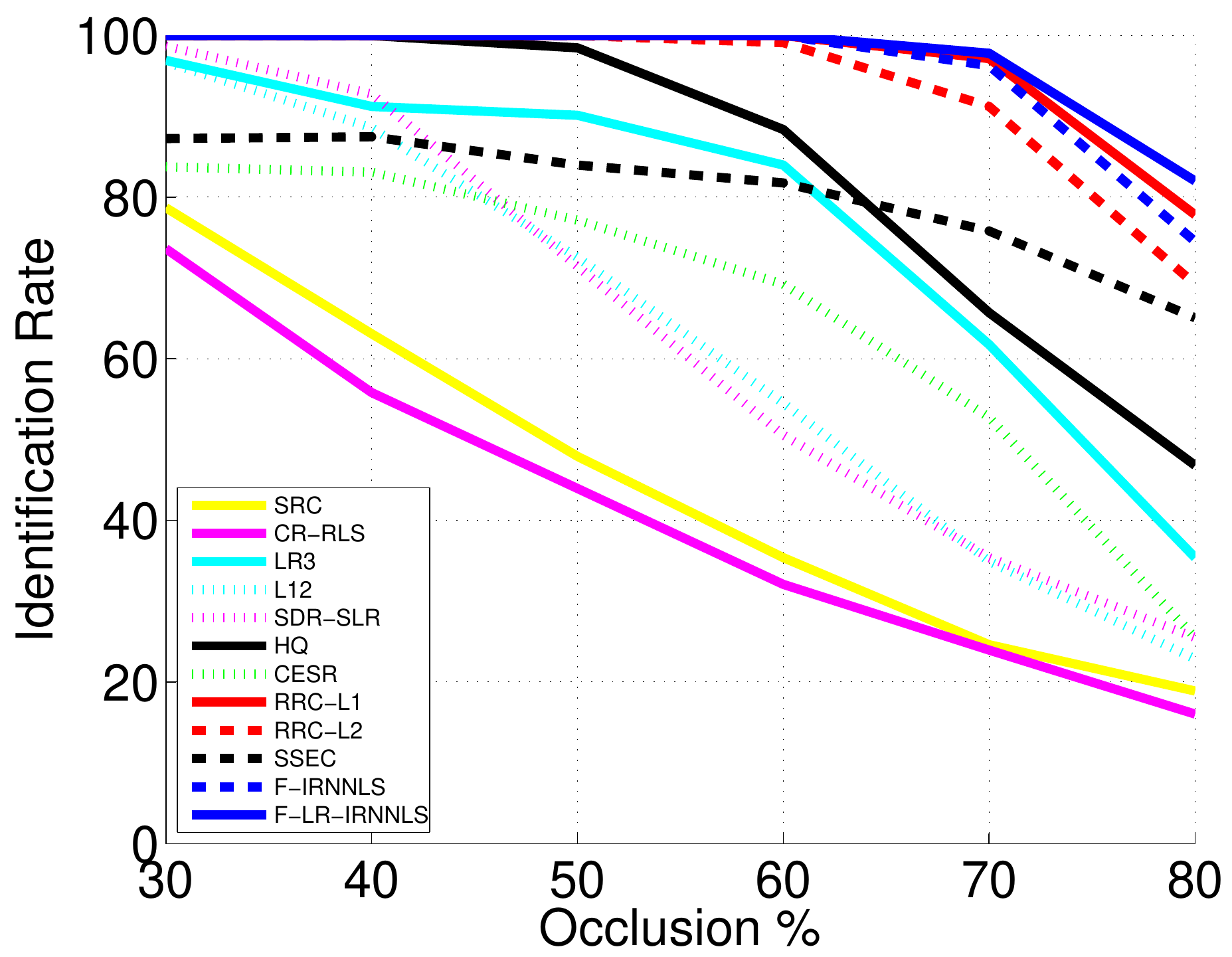}
\caption{ Yale B dataset with dog image}\label{fig:block_dog}
\end{subfigure}%\hspace{0.1cm}
\begin{subfigure}[b]{0.25\textwidth}\includegraphics[width=\textwidth]{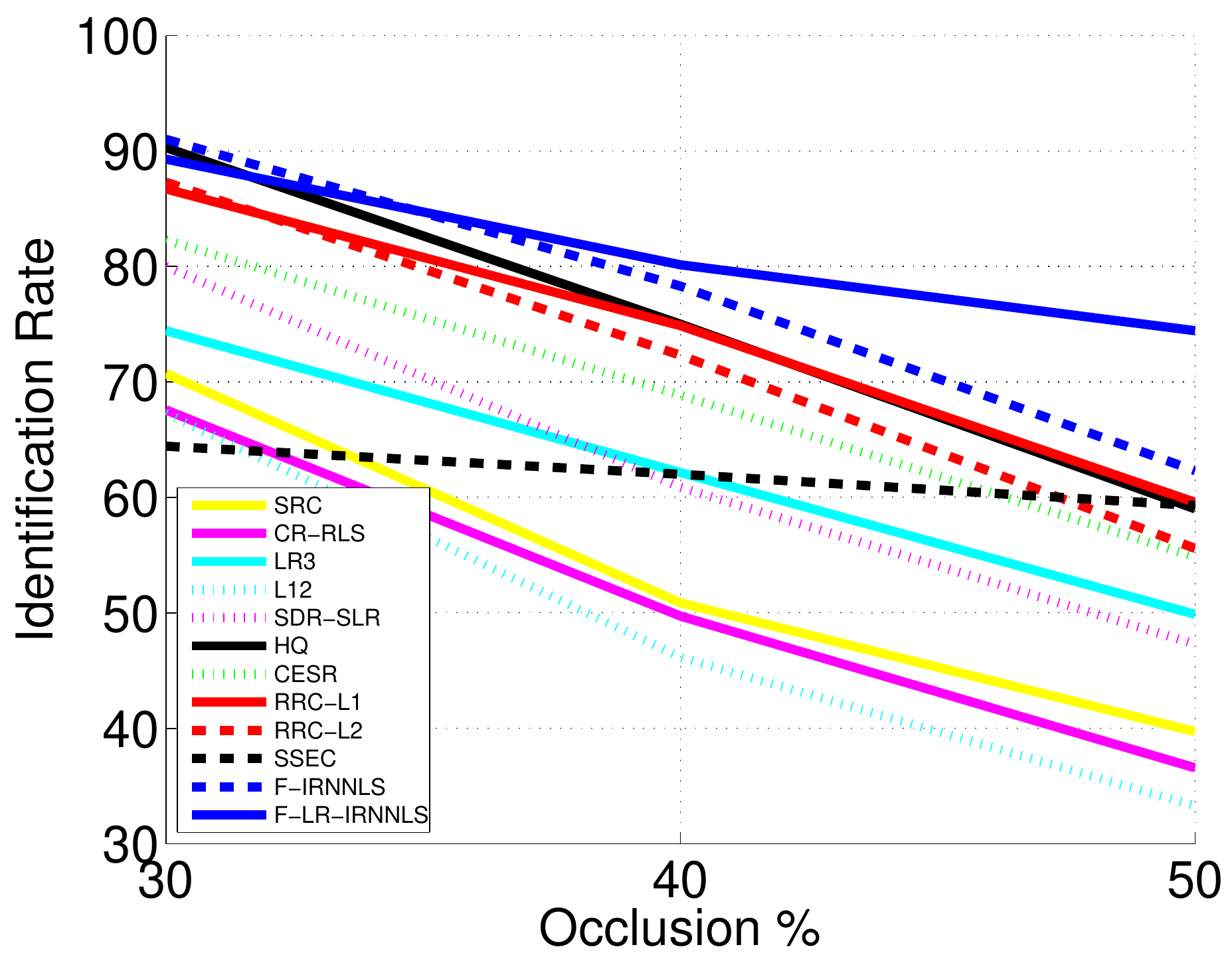}
\caption{AR dataset with baboon image}\label{fig:ar_block}
\end{subfigure}%\hspace{0.1cm}
\caption{Identification Rates on the Extended Yale B and AR under Block Occlusion with Baboon, Vase and Dog.} 
\label{fig:occluded_images_results}
\vspace{-0.3cm}
\end{figure*}

As in \cite{Wright2009,Yang2013,Qian2014}, we chose Subsets 1 and 2 of Extended Yale B for training (in total 719 images) and Subset 3 for testing (in total 455 images). Images were resized to $96\times84$ pixels. We considered three different artificial objects to occlude the test images as shown in Figure~\ref{fig:occluded_images}. For the first object, block occlusion was tested by placing the square baboon image on each test image. The location of the occlusion was randomly chosen and was unknown during training. We considered different sizes of the object such that the face is covered with the occluded object from 30\% to 90\% of its area. Identification rates for the different levels of occlusion are shown in Figure~\ref{fig:occluded_images_results}(a). For the second and third \footnote{ Note that this object (dog) was also tested in work \cite{Li2013}.} non-square and smooth (e.g., without textures in it) objects shown in Figure~\ref{fig:occluded_images}(b) and (c) respectively, block occlusion was tested by randomly placing the objects on each test image. Identification rates are shown in Figure~\ref{fig:occluded_images_results}(b) and \ref{fig:occluded_images_results}(c) respectively for each of the non-square object.

Next, we evaluate the performance to block occlusion in the AR database. We chose the 7 non-occluded AR images per subject (each one with a different face expression) from session 1 for training and the 7 non-occluded images per subject from session 2 for testing. This experiment is more challenging since training faces appear with different expressions. In each test image, we replace a random block with the square baboon image. We resized the images to $60\times43$ pixels. The occlusion ratio increases from 30\% to 50\% and identification rates are shown in Figure~\ref{fig:occluded_images_results}(c).

To examine the robustness of our method on a dataset with more subjects we evaluate its performance to block occlusion in the Multi-PIE database \cite{Gross2010}. The database contains the images of 337 subjects captured in 4 sessions with simultaneous variations in pose, expression, and illumination. In the experiments we used all 249 subjects in Session 1. We used 6 frontal images with 6 illuminations\footnote{Illuminations 0,1,3,4,6,7.} and neutral expression from Session 1 for training, and 10 frontal images\footnote{Illuminations 0,2,4,6,8,10,12,14,16,18.} from Session 4 for testing. In this dataset we also examine the robustness of the methods for different percentages of occlusion under the same experiment and same parameters of the methods. In each test image, we replace a random block with the square baboon image. The percentage of the occlusion was randomly chosen from 30\% to 60\% of the image area. Thus, each test image was chosen to have different percentage of occlusion. Identification rates are shown in Table~\ref{tb:multipie_occlusion}.

From the results with the baboon image we conclude that methods robust to contiguous errors ($\text{LR}^3$, SSEC) performed better than the non-contiguous methods in Yale B. Our F-LR-IRNNLS algorithm outperformed all previous methods overall in all datasets, Yale B, AR and Multi-PIE. SSEC performed well with 80\% occlusion in Yale B but it performed poorly in AR and with lower levels of occlusion. As explained in \cite{Qian2014}, there are no convergence guarantees for SSEC and, perhaps, this explains the unstable results obtained by this method in our experiments. Finally, the performance of non-contiguous error methods, HQ, RRCs and F-IRNNLS significantly dropped by high levels of occlusion. This is due to the fact that these methods cannot handle contiguous variations effectively. In Multi-PIE dataset (Table~\ref{tb:multipie_occlusion} with 6 samples) the F-LR-IRNNLS algorithm performed better than the previous methods. This also suggests that our method performs well under datasets with many subjects and with different occlusion rates. 

From the results with the non-square object and dog image we found out that all methods performed better than when the baboon was used. We attribute this to the fact that the baboon object exhibits a lot of textures and that it looks like a face. Thus, it is much more difficult for the methods to distinguish inliers and outliers. In particular, for the vase image $\text{LR}^3$ and F-LR-IRNNLS achieved similar identification rates at all levels of occlusion. Perhaps, modeling the error to be low-rank was sufficient for the vase image in the Yale B dataset (the weighted norm was unnecessary). Although the identification rates of SSEC were better than those provided by the RRC the method performed significantly worse than F-LR-IRNNLS. SSEC may not be effective in cases where faces are occluded by non-square objects. Finally, as expected the SDR-SLR method performed significantly better than SRC and CR-RLS in all block occlusion experiments. This is due to the fact that the additional intra-class variation dictionary in SDR-SLR captures more effectively some of the variations of the query. %{\color{red}However, since the additional variation dictionary cannot capture all possible occlusions appearing on the test images, SDR-SLR method had lower performance than our F-LR-IRNNLS by a big margin.}}  

The time performance for the block occlusion experiments is presented in Table~\ref{tb:time_performance} (columns 3 and 4)\footnote{In Table~\ref{tb:time_performance} we do not report results from the non-robust methods SRC and CR-RLS since they are both faster than all robust methods. However, they perform poorly in terms of identification rates in all experiments. In this work, our scope is to compare time performance between robust methods.} for Yale B and AR datasets with the baboon image\footnote{Similar time results were obtained for the non-square image.}. A key observation is that while F-IRNNLS achieved identical identification rates with RRC\_L1 and RRC\_L2 it is computationally more efficient by a magnitude. %In addition, F-LR-IRNNLS outperformed RRC\_L1 and RRC\_L2 in running time and had competitive execution times with HQ, CESR, SSEC, however, as discussed previously F-LR-IRNNLS achieved overall higher identification rates.

\subsection{Identification under Expressions \& Face Disguise}

In this experiment we tested our algorithms with face expressions and occlusion with real-world objects in three different scenarios: 1) faces with expressions 2) faces with sunglasses and 3) faces with scarves. The training set consists of the two neutral images (one from each session) from the AR dataset per subject. For the first scenario (face expressions) the 6 images per subject from sessions 1 and 2 with face expressions (smile, anger and scream) were selected for the testing set. For the second scenario (faces with sunglasses) the testing set consisted of the 6 images per subject with sunglasses from sessions 1 and 2. In the third scenario (faces with scarves) the 6 images per subject with scarves from sessions 1 and 2 were chosen for the testing set. The images were resized to $60\times43$ pixels. Identification rates for the three scenarios are shown in Table~\ref{tb:disguise} for the various methods.          

For the face expressions experiment all robust algorithms achieved high performance. A key observation for this experiment is that modeling the error as low-rank does not improve the results since face expression errors do not in general form a contiguous area. 

For the sunglasses experiment we observed that our F-LR-IRNNLS algorithm outperformed previous methods and was able to detect the outliers effectively. In this case modeling the error to be low-rank was adequate. This is due to the fact that the residual image consisted mainly of the sunglasses that made a contiguous error. SSEC performed poorly, perhaps, because the method does not capture well contiguous areas that are not square. A similar conclusion was drawn above with the random block occlusion experiment with a non-square image. 

Results with the scarves experiment demonstrate that all methods robust to contiguous errors performed well, as expected (since the scarf occlusion is contiguous). Our F-LR-IRNNLS method achieved the best performance with 78.83\% identification rate while our non-contiguous F-IRNNLS method achieved only 53.67\%. This result emphasizes the fact that exploiting the spatial correlation in contiguous variations, such as scarves, is beneficial. Time performance is not reported here since all methods run very fast (less than a second) due to the fact that the training dictionary in this experiment was relatively small (200 training samples).

\begin{table*}
\caption{Average run time per test sample on the Extended Yale B and AR datasets under different variations.}
%\vspace{-0.3cm}
\begin{center}
\resizebox{!}{1.5cm}{
  \begin{tabular}{lSSSSSSSSSS}
    \toprule
    \multirow{2}{*}{Dataset} &
      \multicolumn{2}{c}{{\bf Yale Baboon Occl. 60\%}} &
      \multicolumn{2}{c}{{\bf AR Baboon Occl. 50\%}} &
      \multicolumn{2}{c}{{\bf Yale Corruption 90\%}} &
      \multicolumn{2}{c}{{\bf AR Corruption 70\%}} \\      
      & {Accuracy} & {Time} & {Accuracy} & {Time} & {Accuracy} & {Time} & {Accuracy} & {Time} \\
      \midrule
    $\text{LR}^3$ \cite{Qian2014}&88.57\%&0.06s&49.86\%&0.02s&14.29\%&0.06s&n/a&n/a\\
    HQ \cite{He2014} &74.95\%&3.11s&59.00\%&0.99s&42.64\%&1.96s&85.54\%&1.49s\\ 
    CESR \cite{He2011} &41.76\%&0.78s&54.86\%&0.34s&36.04\%&0.99s&85.69\%&0.38s\\ 
    RRC\_L1 \cite{Yang2013} &78.24\%&12.41s&59.57\%&3.95s&{\bf \;\;\;84.40}\%&15.23s&{\bf \;\;\;91.92}\%&29.84s\\
        RRC\_L2 \cite{Yang2013} &81.54\%&10.52s&55.57\%&1.96s&76.26\%&7.40s&88.38\%&10.93s\\
        SSEC \cite{Li2013} &76.70\%&1.58s&59.29\%&0.80s&4.84\%&2.65s&n/a&n/a\\ \hline
         Our F-IRNNLS &80.22\%&1.52s&62.29\%&0.45s&79.78\%&2.84s&91.62\%&2.25s\\ 
         Our F-LR-IRNNLS &{\bf \;\;\;95.82}\%&2.41s&{\bf \;\;\;74.43}\%&0.57s&71.87\%&4.30s&n/a&n/a\\ \hline 
    \bottomrule
  \end{tabular}
}
\end{center}
\label{tb:time_performance}
\vspace{-0.5cm}
\end{table*}

\begin{table}
\caption{Identification Rates (\%) under Face Disguise on the AR database.}
%\vspace{-0.3cm}
\begin{center}
\resizebox{!}{2.1cm}{
\begin{tabular}{c|c|c|c}
\hline
\multirow{1}{*}{Case} 
&Expressions&Sunglasses&Scarves \\
\hline\hline
SRC \cite{Wright2009}&82.33\%&37.17\%&35.17\%\\
CR-RLS \cite{Zhang2011}&81.33\%&33.83\%&39.33\%\\
$\text{LR}^3$ \cite{Qian2014}&79.50\%&80.50\%&77.00\%\\
L12 &85.17\%&40.33\%&43.00\%\\
SDR-SLR \cite{Jiang2015}&92.87\%&36.67\%&32.83\%\\
HQ \cite{He2014}&94.33\%&66.67\%&45.67\%\\ 
CESR \cite{He2011}&93.50\%&66.50\%&17.17\%\\ 
RRC\_L1 \cite{Yang2013}&94.33\%&83.00\%&58.50\%\\
RRC\_L2 \cite{Yang2013}&93.67\%&84.17\%&72.50\%\\
SSEC \cite{Li2013}&56.17\%&70.67\%&{75.00}\%\\ \hline
Our F-IRNNLS &{\bf 94.83}\%&81.50\%&53.67\%\\ 
Our F-LR-IRNNLS &93.00\%&{\bf 89.83}\%&{\bf 78.83}\%  \\ \hline
\end{tabular}
}
\end{center}
\label{tb:disguise}
\vspace{-0.5cm}
\end{table}

\begin{table}
\caption{Identification Rates (\%) and Time Performance (s) under Mixture Noise: Yale B 30\% corruption \& 60\% occlusion, AR 20\% corruption \& 50\% occlusion.}
%\vspace{-0.3cm}
\begin{center}
\resizebox{!}{2.0cm}{
  \begin{tabular}{lSSSSSSSSSS}
    \toprule
    \multirow{2}{*}{Dataset} &
      \multicolumn{2}{c}{{\bf Yale B}} &
      \multicolumn{1}{c}{{\bf AR}} \\      
      & {Accuracy} & {Time} & {Accuracy} \\
      \midrule
SRC \cite{Wright2009}&26.81\%&1.04s&27.86\%\\
CR-RLS \cite{Zhang2011}&14.73\%&0.02s&28.29\%\\
$\text{LR}^3$ \cite{Qian2014}&44.62\%&0.06s&35.43\%\\
SDR-SLR \cite{Jiang2015}&32.75\%&0.50s&40.00\%\\
HQ \cite{He2014} &42.20\%&3.06s&42.00\%\\ 
CESR \cite{He2011} &23.96\%&0.86s&40.86\%\\ 
RRC\_L1 \cite{Yang2013} &43.08\%&14.58s&47.00\%\\
RRC\_L2 \cite{Yang2013} &41.54\%&9.65s&38.14\%\\
SSEC \cite{Li2013} &14.95\%&1.72s&11.86\%\\ \hline
Our F-IRNNLS &45.27\%&1.45s&49.00\%\\ 
Our F-LR-IRNNLS &{\bf \;\;\;63.08}\%&6.33s&{\bf \;\; 57.29}\%\\ \hline
    \bottomrule
  \end{tabular}
}
\end{center}
\label{tb:mixture}
\vspace{-0.9cm}
\end{table}

\subsection{Identification under Mixture Noise}

In this experiment we evaluate the performance of our algorithm for the case of mixture noise. In this case, both pixel corruption and block occlusion degraded the testing images. An example image with this degradation is shown in Figure~\ref{fig:qual_results}. This experiment was conducted with two datasets, Extended Yale B and AR. Similarly to the previous Extended Yale B settings, Subsets 1 and 2 of Extended Yale B were used for training and Subset 3 was used for testing. With the AR dataset we chose the 700 non-occluded AR images for training from session 1 and the 700 non-occluded images for testing from session 2. In both datasets, for each testing image a percentage of randomly chosen pixels was corrupted. Corruption was performed by replacing those pixel values with independent and identically distributed samples from a uniform distribution between [$0$, $255$]. Then, we placed the baboon square image on each corrupted test image. In Yale B dataset we performed this experiment with 30\% pixel corruption and 60\% occlusion. With the AR dataset, experiments were conducted with 20\% pixel corruption and 70\% occlusion. Identification rates are shown in Table~\ref{tb:mixture} for the various methods.

F-LR-IRNNLS outperformed all previous methods which indicates that in the mixture noise case, our two error constraints capture the error term effectively. SSEC performed poorly due to the presence of pixel corruption. RRC\_L1, RRC\_L2 and HQ were robust to pixel corruption, however, their performance remained low since they were not effective on describing the occlusion part. Our F-LR-IRNNLS had a good balance on detecting the corrupted pixels and capturing the occlusion part with the employment of the weighted and nuclear norms. However, although F-LR-IRNNLS achieved significantly higher performance than the previous methods, the actual accuracy was relative low with 63.08\% in YaleB and 57.29\% in AR. The result may indicate that in mixture of noises further investigation about modeling the error is required.    

Execution times in this case are reported in Table~\ref{tb:mixture}. A key observation is that F-IRNNLS is by an order of magnitude faster than RRC\_L1 and RRC\_L2. F-LR-IRNNLS was faster than RRCs and slower than $\text{LR}^3$. However, $\text{LR}^3$ achieved significantly lower identification rates.

\begin{table}
\caption{Identification Rates on the Multi-PIE under Illumination.}
%\vspace{-0.3cm}
\begin{center}
\resizebox{!}{2.0cm}{
  \begin{tabular}{lSSSSSSSSSS}
    \toprule
    \multirow{2}{*}{Sessions} &
      \multicolumn{1}{c}{{\bf Session 2}} &
      \multicolumn{1}{c}{{\bf Session 3}} &
      \multicolumn{2}{c}{{\bf Session 4}} \\   
      & {Accuracy} & {Accuracy} & {Accuracy} & {Time} \\
      \midrule
   SRC \cite{Wright2009}&95.48\%&92.13\%&95.71\%&0.63s\\
CR-RLS \cite{Zhang2011}&95.30\%&90.56\%&94.46\%&0.02s\\
$\text{LR}^3$ \cite{Qian2014}&90.00\%&83.44\%&87.89\%&0.04s\\
L12 &93.92\%&88.69\%&92.91\%&0.04s\\
HQ \cite{He2014}&95.84\%&94.63\%&97.14\%&2.21s\\ 
CESR \cite{He2011}&94.64\%&92.94\%&96.06\%&1.94s\\ 
RRC\_L1 \cite{Yang2013}&95.84\%&95.13\%&97.14\%&6.88s\\
RRC\_L2 \cite{Yang2013}& 97.11\%&94.19\%&97.37\%&31.11s\\
SSEC \cite{Li2013}&86.75\%&76.13\%&83.09\%&5.40s\\ \hline
Our F-IRNNLS &{\bf \;\; 97.29}\%&{\bf \;\;\;96.00}\%&{\bf \;\; 98.00}\%&1.05s\\ 
Our F-LR-IRNNLS &96.87\%&94.13\%&97.83\%&1.83s\\ \hline
    \bottomrule
  \end{tabular}
}
\end{center}
\label{tb:multipie}
\vspace{-0.5cm}
\end{table}

\subsection{Identification under Illumination}

Experiments with variations in illumination were conducted on the Multi-PIE dataset. As in the block occlusion experiments, we used all 249 subjects in Session 1. As in \cite{Zhang2011}, we used 14 frontal images with 14 illuminations\footnote{Illuminations 0,1,3,4,6,7,8,11,13,14,16,17,18,19.} and neutral expression from Session 1 for training, and 10 frontal images\footnote{Illuminations 0,2,4,6,8,10,12,14,16,18.} from Session 4 for testing. Identification rates are shown in Table~\ref{tb:multipie} for the various methods.

Our first observation is that all methods achieved high identification rates. Simple SRC approaches performed well while robust methods only slightly improved the results. The reason with respect to our method is that for illumination variations modeling the error image as low-rank does not hold in this case. Similar observations can also be deduced from results of the $\text{LR}^3$ method in this case.   

%The reason is that in this experiment lighting intra-class variations exist in training samples and thus, the illumination variations are sufficiently reconstructed by the linear combination of the available training images.

With respect to time performance, our algorithm outperforms the previous robust methods. In particular the execution time in our approaches is around 1 second per test image while for RRC\_L2 is around 30 seconds. Notice that although this was an experiment with a large training dictionary, our method retains very low running time. 

%We also used the cropped face images with dimensions $50 \times 41$ pixels

\begin{figure}[!t]
\centering          
\begin{subfigure}[b]{0.24\textwidth}\includegraphics[width=\textwidth]{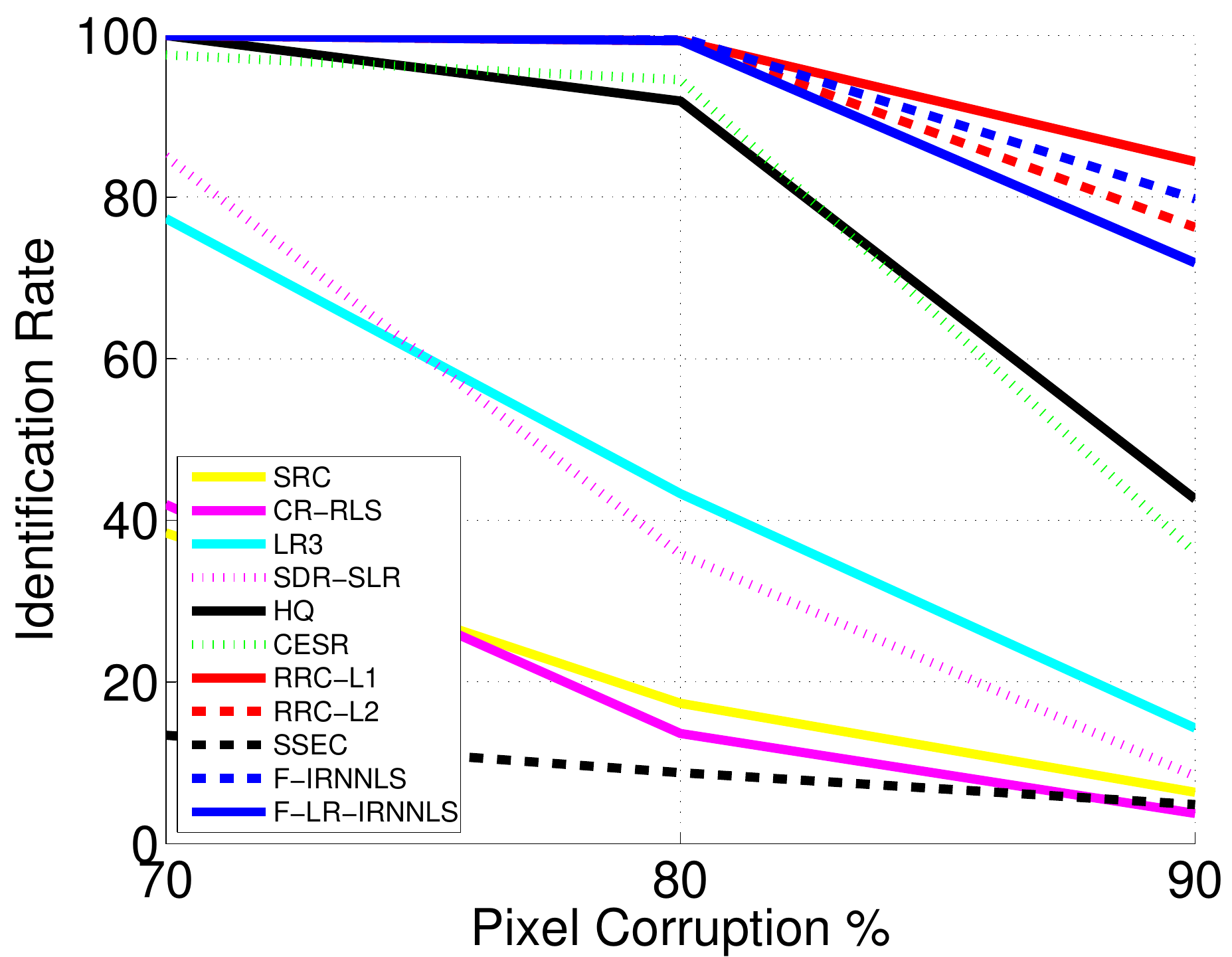}
\caption{Yale B dataset}\label{fig:corruption}
\end{subfigure}%\hspace{0.1cm}
\begin{subfigure}[b]{0.24\textwidth}\includegraphics[width=\textwidth]{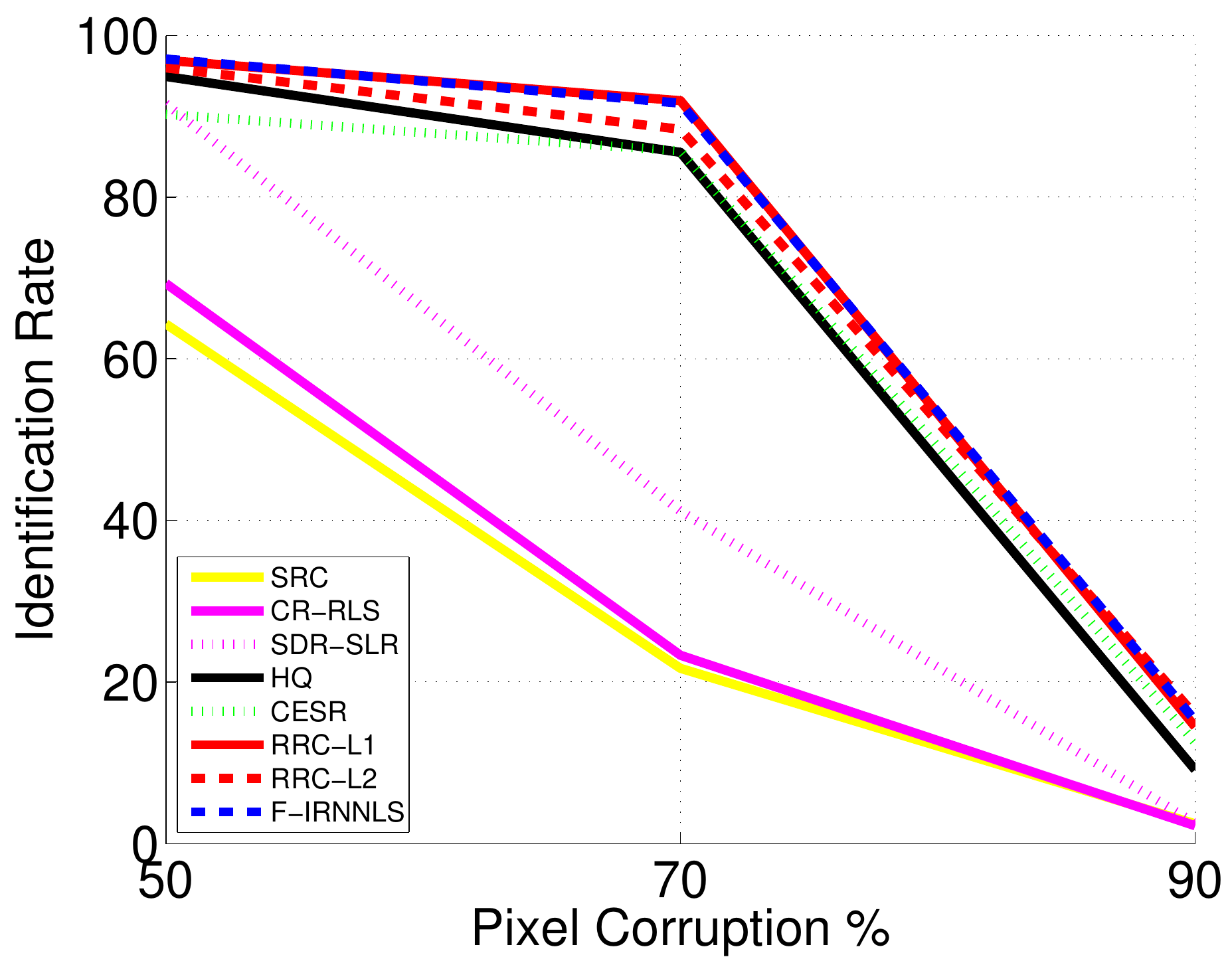}
\caption{AR dataset}\label{fig:ar_corruption}
\end{subfigure}%\hspace{0.1cm}
\caption{Identification Rates on the Extended Yale B and AR datasets under Pixel Corruptions.} 
\label{fig:corrupted_images}
%\vspace{-0.5cm}
\end{figure}

\begin{table}
\caption{ Identification Rates (\%) in the unconstrained dataset LFW-a in the left column. In the right column we demonstrate identification rates (\%) in Multi-PIE dataset under occluded training samples.}
%\vspace{-0.3cm}
\begin{center}
\resizebox{!}{2.2cm}{
  \begin{tabular}{lSSSSSSSSSS}
    \toprule
    \multirow{1}{*}{Dataset} &
      \multicolumn{1}{c}{{\bf LFW-a}} &
      \multicolumn{1}{c}{{\bf Multi-PIE}}     
       \\        
      \midrule
SRC \cite{Wright2009} & 68.35\%&14.86\%\\
CR-RLS \cite{Zhang2011}& 64.68\%&6.91\%\\
$\text{LR}^3$ \cite{Qian2014}& 62.53\%&6.00\%\\
SDR-SLR \cite{Jiang2015} & 76.20\%&39.03\%\\ 
HQ \cite{He2014} & 71.00\%&17.83\%\\ 
CESR \cite{He2011} & 59.87\%&11.43\%\\ 
RRC\_L1 \cite{Yang2013} & 63.42\% &23.20\%\\
RRC\_L2 \cite{Yang2013} & 72.53\% &37.66\%\\
SSEC \cite{Li2013} & 32.41 \% &44.14\%\\ \hline
Our F-IRNNLS &72.91\% &23.89\%\\ 
Our F-LR-IRNNLS & 74.81\% &44.06\%\\
Our F-LR-IRNNLS (SLR) & \;\; {\bf 77.47}\% &{\bf \;\; 47.71}\%\\ \hline
    \bottomrule
  \end{tabular}
}
\end{center}
\label{tb:lfw}
\vspace{-0.5cm}
\end{table}

\subsection{Identification under Pixel Corruptions}

Experiments under pixel corruption were conducted on two datasets: Extended Yale B and AR.

As in \cite{Wright2009,Yang2013} we used the non-occluded faces of Subsets 1 and 2 of the Extended Yale B for training and Subset 3 for testing. Images were resized to $96\times84$ pixels. In the AR dataset, to make the experiment more challenging we chose to use occluded training and testing images. AR has 100 different subjects and for each subject the 13 images from session 1 (7 non-occluded images, 3 images with sunglasses, and 3 images with scarves) were used for training and the 13 images from session 2 for testing. Images were resized to $60\times43$ pixels.

For each test image in both datasets, a percentage of randomly chosen pixels was corrupted by replacing those pixel values with independent and identically distributed values from a uniform distribution between [$0$, $255$]. The percentage of corrupted pixels was varied between $50$ percent and $90$ percent. Identification rates are shown in Figure~\ref{fig:corrupted_images} for the various methods.

Figure~\ref{fig:corrupted_images}(a) illustrates that the robust non-contiguous methods RRC\_L1 and F-IRNNLS achieved the best performance with over 80\% accuracy in 90\% pixel corruption. Methods able to handle contiguous errors such as SSEC, $\text{LR}^3$ and F-LR-IRNNLS performed poorly. We attribute this to the fact that pixel corruption is not a contiguous variation and modeling the error to have contiguous structure was inadequate. To that extend, with the AR dataset we decided to report results only on methods that handle non-contiguous errors as shown in Figure~\ref{fig:corrupted_images}(b). In this dataset the accuracy is low in 90\% corruption for all methods. As explained earlier, this was a more challenging experiment with a large number of testing images consisting of faces with pixel corruption on top of occlusion.

With respect to execution time, our F-IRNNLS method outperformed RRC\_L1 and RRC\_L2 in both Yale B and AR datasets, as shown in Table~\ref{tb:time_performance} (columns 1 and 2). To emphasize the difference in performance, the execution time of F-IRNNLS in AR for 70\% pixel corruption was about 2 seconds. In RRC\_L1 and RRC\_L2 was 29.84 and 10.93 seconds, respectively. Time performance of our algorithms was comparable to CESR and HQ (additive form) and worse than $\text{LR}^3$. However, these methods obtained lower identification rates. 

\begin{figure}[!t]
\centering
\begin{subfigure}[b]{0.42\textwidth}\includegraphics[width=\textwidth]{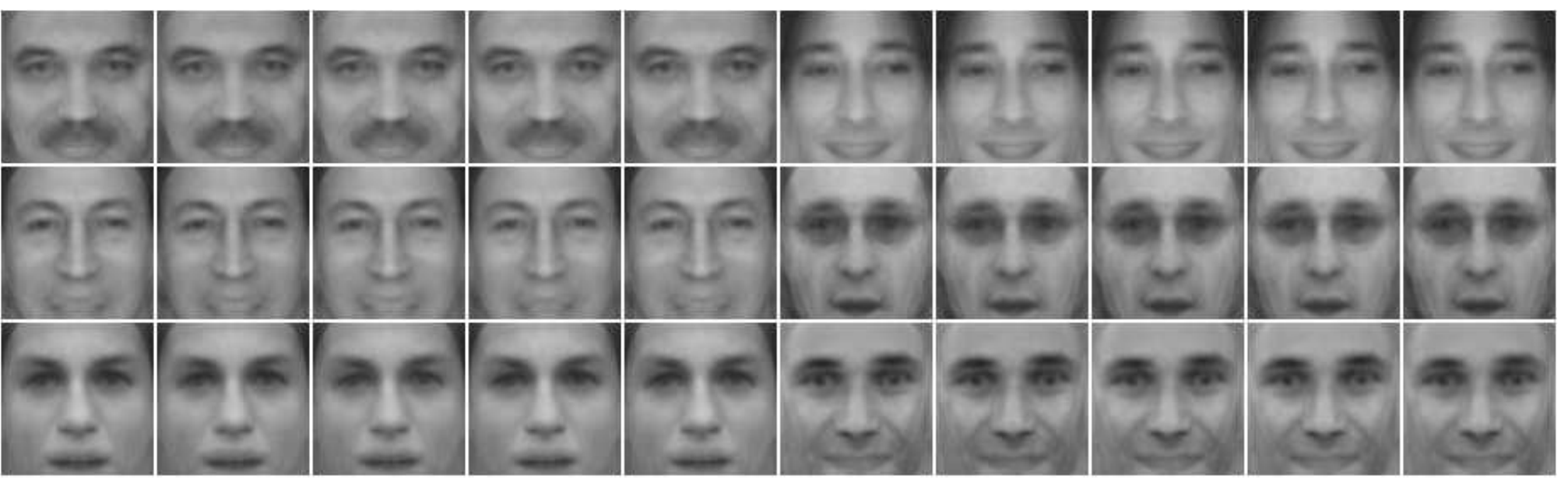}
\caption{ Training sample images from the class-specific dictionary ${\bf A}$.}
\end{subfigure}\\%\hspace{0.1cm}
\begin{subfigure}[b]{0.42\textwidth}\includegraphics[width=\textwidth]{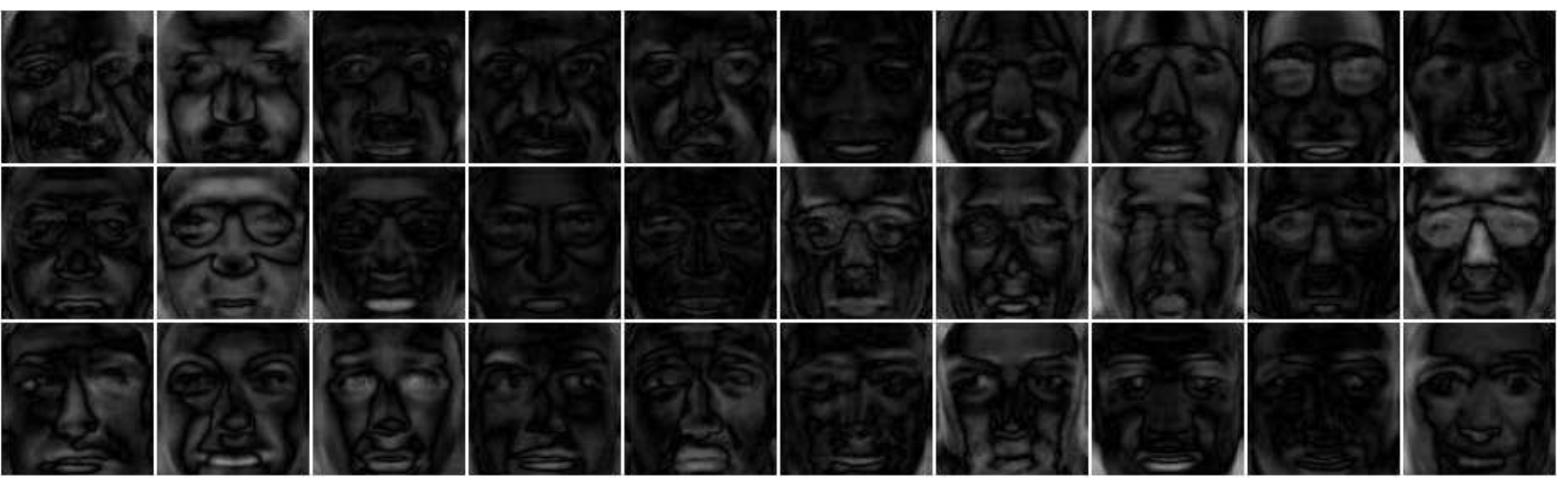}
\caption{ Training sample images from the variation dictionary ${\bf B}$.}
\end{subfigure}\\%\hspace{0.1cm}
\caption{ Sample images from the LFW-a dataset and the SDR-SLR decomposition applied to the dataset.} 
\label{fig:lfw_samples}
\vspace{-0.5cm}
\end{figure}

\begin{figure*}[!t]
%\begin{table}
     \centering
	\includegraphics[scale=0.37]{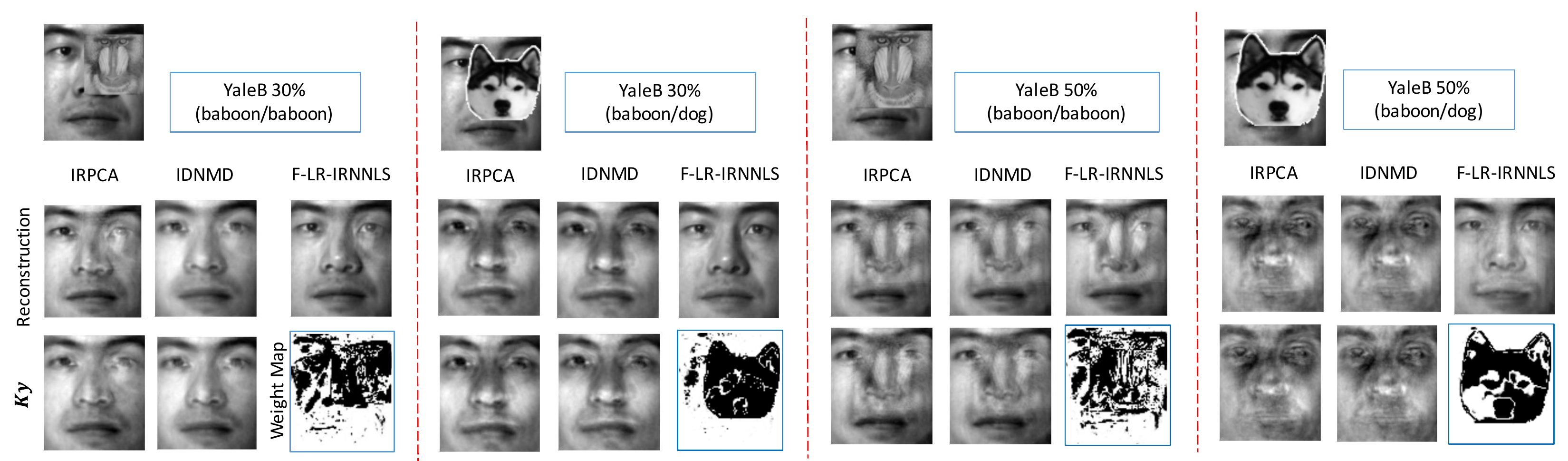}
\begin{center}
\resizebox{!}{1.3cm}{
  \begin{tabular}{lSSSSSSSSSS}
    \toprule
    \multirow{2}{*}{Dataset} &
      \multicolumn{1}{c}{{\bf YaleB 30\%}} & 
      \multicolumn{1}{c}{{\bf YaleB 50\%}} &
      \multicolumn{1}{c}{{\bf YaleB 60\% }} &
      \multicolumn{1}{c}{{\bf YaleB 50\% }} &
      \multicolumn{1}{c}{{\bf AR 50\%}} &
      \multicolumn{1}{c}{{\bf AR 50\%}} &
      \multicolumn{1}{c}{{\bf LFW-a}} \\
      & { baboon/baboon} & {baboon/baboon} & {baboon/baboon} & {baboon/dog} & {baboon/baboon} & {baboon/dog} & {-} \\
      \midrule
      SRC \cite{Wright2009} & 95.16\% & 67.91\% & 49.23\% & 37.58\% & 34.29\% &21.86\% & 68.35\%   \\ 
IRPCA \cite{Bao2012} & 94.73\% & 74.95\% & 57.36\% & 30.33\%  & 42.43\% & 26.00\% & 69.87\% \\
    IDNMD \cite{Zhang2015} & 95.60\% & 75.38\%  & 60.44\% & 32.09\%  & 43.29\% &22.86\%  & 69.37\% \\
    $\text{LR}^3$ \cite{Qian2014} & 94.95\% & 73.63\% & 53.85\% & 46.81\% & 46.81\% & 36.14\% & 62.53\% \\ \hline 
    F-LR-IRNNLS & { \;\;\bf 99.12\%} & { \;\;\;\bf 82.41\%} & { \;\; \bf 62.64\%} & { \;\;\bf 90.93\%} & { \;\; \bf 72.43\%} &{ \;\; \bf 75.86\%} & { \; \bf 74.81\%} \\ 
    \bottomrule
  \end{tabular}
}
\end{center}

\caption{Comparisons between our method and inductive approaches. Figure: recovery results of three methods. The first row shows reconstructed faces ${\bf Ta}$ provided by the classifier while the second row visualizes the recovered test sample ${\bf Ky}$ from the inductive methods and weight map estimations of our method. Table: of identification rates  of the methods. We denote the different training and testing occlusion scenarios as following; ``baboon/baboon" is that training and testing samples are occluded with the baboon image. ``baboon/dog" is that training samples are occluded with the baboon image and testing images are occluded with the dog image.}

\label{tb:inductive_occlusion}
\vspace{-0.5cm}
%\end{table}
\end{figure*}

\subsection{Identification under Unconstrained Environment and Occluded Testing and Training Samples}

\subsubsection{Identification under Unconstrained Environment}

Thus far, we have assumed that training samples are ``clean" frontal aligned views and without large variations of the same identity. In an unconstrained environment this assumption does not hold and often face images of the same identity exhibit large variations in pose, illumination, expression and occlusion. Furthermore, testing images may not contain the same variations and occlusions as the training images. 

To examine the robustness of our method on an unconstrained environement we evaluate its performance in the LFW database. The dataset contains images of 5,749 different subjects and in this work we used the LFW-a \cite{Wolf2010}, which is an aligned version of LFW based on commercial face alignment software. We used the subjects that include no less than ten samples and we constructed a dataset with 158 subjects from LFW-a. For each subject, we randomly chose 5 samples for training (resulting in a dictionary of 790 faces) and 5 samples for testing. The images were resized to $90 \times 90$.

To deal with such environment we utilize the SDR-SLR algorithm as a pre-processing step as explained in Section~\ref{sec:corrupted_training}. Sample images from dictionaries ${\bf A}$ and ${\bf B}$ estimated on LFW-a dataset are illustrated in Figure~\ref{fig:lfw_samples}(a) and (b). More specifically, in Figure~\ref{fig:lfw_samples}(b) images cover variations which are not class-specific and are used to represent complex variations of the query. These variations may be represented by the component ${\bf B}$ using images that do not belong to the same identity of the test image. Any remaining variations that cannon be described by ${\bf B}$ are captured by the term ${\bf e}$.  

Identification rates for the LFW dataset are shown in Table~\ref{tb:lfw}(left column) for the various methods. Our first observation is that the method SDR-SLR achieved better performance than our methods F-IRNNLS and F-LR-IRNNLS. This is expected as in this experiment faces exhibit uncontrolled variations such as pose and expression which are learnt from the training data utilized in SDR-SLR. However, when the method SDR-SLR is combined with our F-LR-IRNNLS method denoted as F-LR-IRNNLS (SLR), performance is improved. This is due to the fact that there are some remaining variations of the query such as occlusion that cannot be learnt from the training data. Our method is able to represent these remaining variations by modeling the representation error as low-rank and fitting to the error a distribution described by a tailored loss function.

\subsubsection{Identification under Occluded Testing and Training Samples}

In order to further investigate the scenario where we are given corrupted testing and training data in a constrained environment this time, 

We also conducted an experiment on the Multi-PIE dataset in the scenario that corrupted testing and training data are provided in a constrained environment. To simulate the corrupted (occluded) training data we considered the same training and test sets as in Multi-PIE block-occlusion experiment described above. In particular, we used the 6 frontal images with 6 illuminations and neutral expression from Session 1 for training. For each of the 249 subjects we randomly chose half of the training images to be occluded. In each image we replace a random block with the square baboon image with occlusion chosen randomly from 30\% to 60\%. We chose the 10 frontal images from Session 4 for testing. In each test image, we replace a random block with the square baboon image and the occlusion was randomly chosen from 30\% to 60\%. Identification rates are shown in Table~\ref{tb:lfw}(right column).

From the results we observe that as expected SDR-SLR perform way better than the SRC and CR-RLS. However, not all block-occlusions are sufficiently covered by ${\bf B}$. This is due to the fact that occlusion appears in random places and sizes in the query as well as in training data. Therefore, it might be very unlikely that the occlusion on the query and training images will be of the same type. Thus, when SDR-SLR method is combined with the F-LR-IRNNLS algorithm the best accuracy is reported since the proposed modeling of the error term is robust to handle occlusions of the query. Finally, we observe that our F-LR-IRNNLS algorithm outperform the other approaches even when SDR-SLR is not utilized. The reason may be that our approach chooses the non-occluded training samples to represent the query since occlusion is effectively captured by the representation error image ${\bf e}$.

\subsubsection{Comparison with inductive methods}

In this section we compare our method with two inductive methods, namely IRPCA \cite{Bao2012} and IDNMD \cite{Zhang2015}. Both methods are able to handle new data meaning that given a new sample an underlying learnt projection matrix ${\bf K}$ can be used to efficiently remove corruptions and occlusions. A test face is recovered from occlusions by computing ${\bf Ky}$. Then, the ``clean" test face ${\bf Ky}$ is provided as an input to a classifier to identify the subject. In this work we use the SRC classifier \cite{Wright2009} for these methods in order to make direct comparisons with our method \footnote{We report results with names IRPCA and IDNMD to denote IRPCA combined with SRC and IDNMD combined with SRC classifier, respectively.}. Also, in this case we do not perform any pre-processing step such as SDR-SLR to clean the training data for identification in our and other methods to make fair comparisons.

There are two main differences between IRPCA, IDNMD and our method; i) Our method does not require the same class of occlusions to be present in the training and test data while inductive methods do to perform well. ii) Our method describes the error image by using two metrics, namely weighting and nuclear norms.

% which is shown in our previous experiments to be more effective than than characterizing it only as either low-rank (see $\text{LR}^3$ and IDNMD) or sparse (IRPCA) for occlusions.	 

%The error in the face image for our method is characterized by the two metrics described in Section~\ref{sec:firc} which is shown in previous experiments to be more effective than characterizing it only as either low-rank (see $\text{LR}^3$ and IDNMD) or sparse (IRPCA) for occlusions.	

%Quantitative results are given in table~\ref{tb:inductive_occlusion}) and qualitative results are shown in figure~\ref{fig:inductive_errors}.

\begin{figure}[!t]
     \centering
	\includegraphics[scale=0.4]{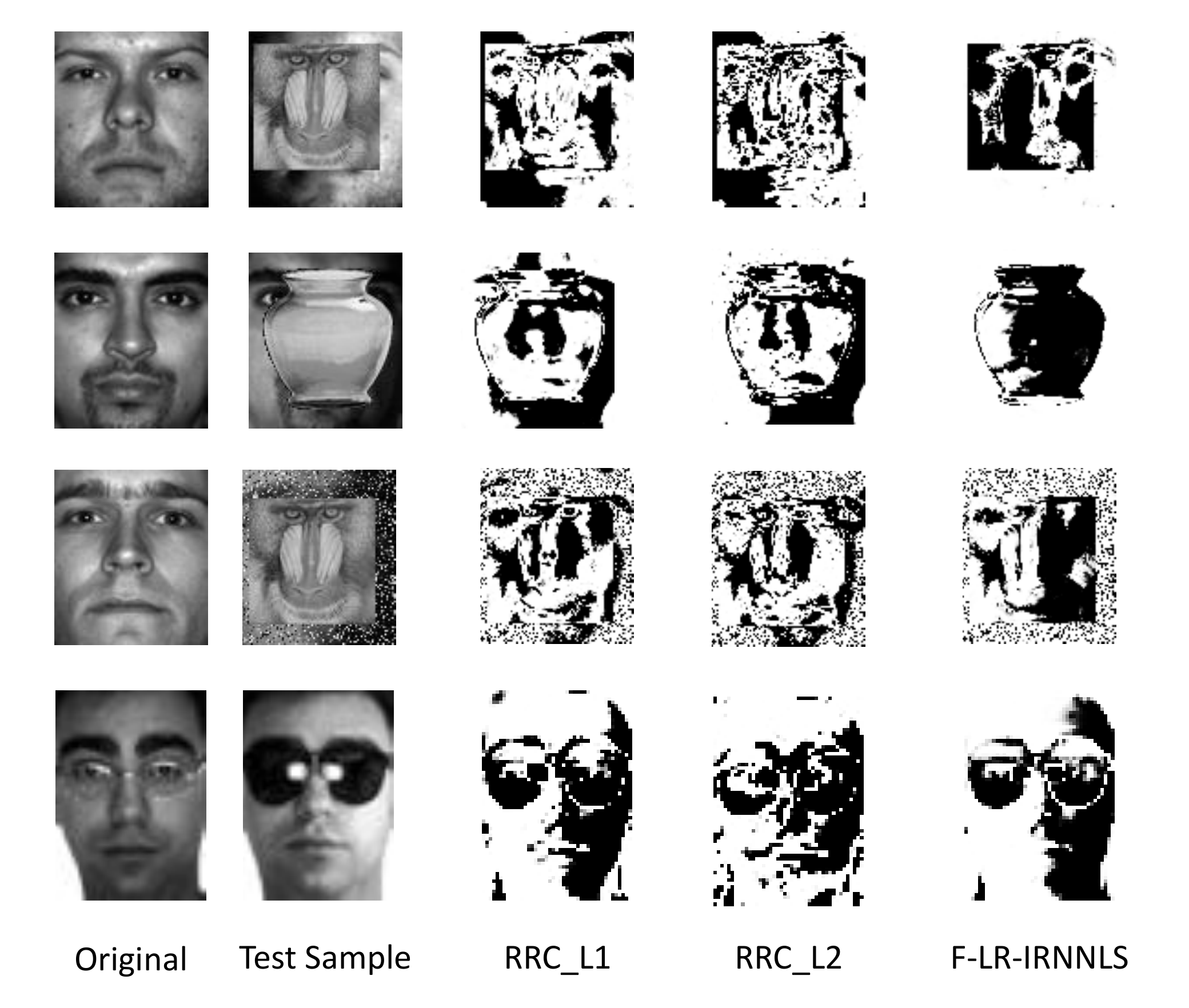}\\
     %\vspace{-0.3cm}
    \caption{Estimated weight maps for three iterative reweighted coding methods. The weight maps of our F-LR-IRNNLS method capture the outlier object of interest more accurately. With the other methods, a number of inlier pixels are detected as outliers.     }
     %\vspace{0.0cm}
     \label{fig:weight_results}
\vspace{-0.5cm}
\end{figure} 

To examine the robustness of our method against the inductive methods we conducted the following experiments:

First we investigate the performance on the YaleB dataset. We chose Subsets 1 and 2 of Extended Yale B for training and Subset 3 for testing. As in \cite{Zhang2015}, 15\% of training images were occluded with the baboon image in random places. Three occlusion rates, 30\%, 50\% and 60\% were considered. For testing images we considered two different scenarios; 1) test images were occluded with the baboon image randomly placed (different random places than in the training set) 2) test images were occluded with the dog image (training samples were still occluded with the baboon image) to explore whether inductive methods can handle occlusions that are not present in the training procedure. From the results in Figure~\ref{tb:inductive_occlusion} (table) for YaleB dataset we deduce the following findings:

As expected, inductive methods combined with SRC perform better than standard SRC methods when the same occlusion type (baboon) is present in training and testing images (second and third columns in Figure~\ref{tb:inductive_occlusion}). For, low-percentage of occlusion (30\%) all methods perform well. However, when occlusion is different in training and testing images (baboon/dog) as shown in Figure~\ref{tb:inductive_occlusion} (fourth column), SRC performs very similarly to inductive methods. This demonstrates that when a specific occlusion object is involved in the test but not the training image, IRPCA and IDNMD may not perform better than SRC. 

This is further explained in Figure~\ref{tb:inductive_occlusion} (figure). First, we observe from this figure that when a low-percentage occlusion is present (30\%), inductive methods can recover the test face (see ${\bf Ky}$) well, and as a consequence can perform well in identification. However, when the occlusion percentage is 50\%, the face recovery by the inductive methods becomes more noisy. The face becomes a lot more noisy, when the testing occlusion (dog) is different from the occlusion used during training (baboon). From the visualizations in Figure~\ref{tb:inductive_occlusion} we also observe that in our method the occluded object is well captured in all cases (see weight maps).

%, since i) no knowledge about the occlusion in test sample is required ii) the two error metrics introduced earlier effectively describe the high percentage of occlusion. 

IDNMD performs slightly better than IRPCA since it models each occlusion-induced image as low-rank during training. Similar results were reported in \cite{Zhang2015}. For a more effective comparison, we also report results in Figure~\ref{tb:inductive_occlusion} for $\text{LR}^3$ \cite{Qian2014}, since $\text{LR}^3$ describes the error as low-rank too. It was expected that IDNMD and $\text{LR}^3$ would have similar performance when training and testing images have similar occlusions. However, when occlusions are different in training and testing images, $\text{LR}^3$ performs better. Our method outperforms significantly both IDNMD and $\text{LR}^3$ methods. 

%Overall, our method outperform previous methods especially in high-percentages of occlusion and when occlusion type is different in training and testing faces.     

The second experiment was conducted on the AR database. We chose the 7 neutral AR images per subject from session 1 for training, where 15\% of those were occluded with the baboon image in random places (occlusion percentage is 50\%). The 7 neutral images per subject from session 2 were chosen for testing. In each test image, we replaced a random block with the dog image. Identification rates are shown in Figure~\ref{tb:inductive_occlusion} (fifth and sixth columns). Results are consistent with the YaleB findings. SRC, IDNMD and IRPCA perform very similarly in fifth column of Figure~\ref{tb:inductive_occlusion} (since different occlusion is presence in training and testing faces) although the identification rates are really low. Our method again outperforms previous methods significantly.  

Finally, we investigate the performance of the algorithms in an unconstrained environment utilizing the LFW database. The experimental settings are similar to the experiment showed in Table~\ref{tb:lfw}. As with previous experiments, from the results we observe that our proposed method outperforms previous approaches as the error image is described more effectively with the two metrics, weighting and nuclear norms.

\begin{table}
\caption{Identification Rates on the Multi-PIE under block occlusion and few training samples.}
%\vspace{-0.3cm}
\begin{center}
\resizebox{!}{2.0cm}{
  \begin{tabular}{lSSSSSSSSSS}
    \toprule
    \multirow{1}{*}{Samples} &
      \multicolumn{1}{c}{{\bf 2 Samples}} &
      \multicolumn{1}{c}{{\bf 4 Samples}} &
      \multicolumn{1}{c}{{\bf 6 Samples}} \\      
      \midrule
   SRC \cite{Wright2009}&9.37\%&13.31\%&22.47\%\\
CR-RLS \cite{Zhang2011}&5.43\%&8.40\%&18.34\%\\
$\text{LR}^3$ \cite{Qian2014}&28.51\%&29.71\%&37.37\%\\
SDR-SLR \cite{Jiang2015} &10.17\%&21.77\%&33.83\%\\
HQ \cite{He2014}&22.91\%&38.23\%&49.31\%\\ 
CESR \cite{He2011}&11.54\%&25.31\%&42.11\%\\ 
RRC\_L1 \cite{Yang2013}&22.11\%&43.66\%&64.46\%\\
RRC\_L2 \cite{Yang2013}&28.86\%&44.51\%&52.86\%\\
SSEC \cite{Li2013}&28.23\%&47.77\%&65.43\%\\ \hline
Our F-IRNNLS &22.00\%&48.86\%&64.74\%\\ 
Our F-LR-IRNNLS &{\bf \;\; 34.46}\%&{\bf \;\; 52.69}\%&{\bf \;\; 71.09}\%\\ \hline
    \bottomrule
  \end{tabular}
}
\end{center}
\label{tb:multipie_occlusion}
\vspace{-0.5cm}
\end{table}

\subsection{Identification under Few Training Samples}

To examine the robustness of our method under few training samples per subject we conducted experiments on the Multi-PIE database \cite{Gross2010} with uncorrupted training data. As in the experiments under occlusion, we used 6 frontal images with 6 illuminations and neutral expression from Session 1 for training, and 10 frontal images\footnote{Illuminations 0,2,4,6,8,10,12,14,16,18.} from Session 4 for testing. Then, we randomly selected 2 or 4 samples per subject to perform  experiments under fewer training examples. In each test image, we replace a random block with the square baboon image, where each block randomly covered between 30\% and 60\% of the image area.

From the results in Table~\ref{tb:multipie_occlusion}, we deduce that as expected all methods perform worse when fewer training samples are available. However, our F-LR-IRNNLS algorithm achieved the best performance in 2 and 4 samples. SSEC achieved the second best performance while $\text{LR}^3$ performed similarly in 2 samples with the SSEC and RRC\_L2. As expected, the SDR-SLR method performed significantly better than SRC and CR-RLS since the additional intra-class variation dictionary alleviates the issue with the limited samples per person. However, since the additional variation dictionary does not capture the unknown occlusion appearing on the test images, SDR-SLR method had lower performance than our F-LR-IRNNLS by a big margin. 

Overall our proposed method achieved higher or competitive identification rates across all experiments. In addition, our method incurred lower computational costs than the previous algorithms. In some cases the execution time was lower by an order of magnitude than the second best algorithm overall (e.g., RRC\_L1, RRC\_L2). Furthermore, our F-LR-IRNNLS and F-LR-IRNNLS (SLR) algorithm outperformed with respect to identification rates all previous robust sparse representation-based methods on images with contiguous errors in all scenarios (uncorrupted training data, occluded testing and training data and unconstrained environment).

\begin{figure}[!t]
\centering
\includegraphics[scale=0.40]{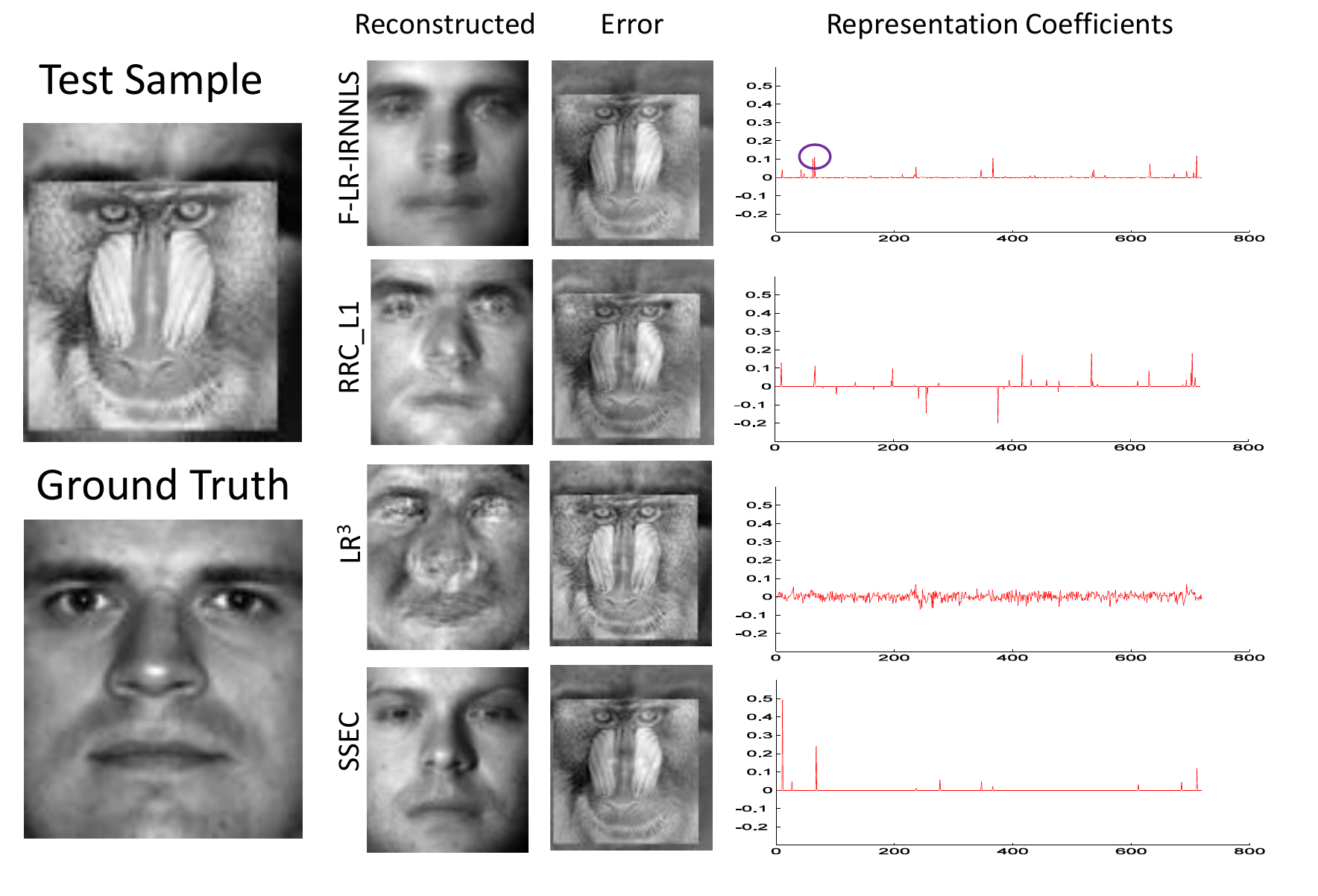}\\
\caption{Reconstruction results in 70\% block occlusion.} 
\label{fig:qualitative_results_b}
\vspace{-0.3cm}
\end{figure}

\begin{table*}
\caption{Comparison of identification rates and times between RRCs and our Algorithms under different regularizations of the coefficients.}
\begin{center}
\resizebox{!}{1.5cm}{
  \begin{tabular}{lSSSSSSSSSS}
    \toprule
    \multirow{2}{*}{Dataset} &
      \multicolumn{2}{c}{{\bf Illumination}} &
      \multicolumn{2}{c}{{\bf Occlusion 60\% (Y)}} &
      \multicolumn{2}{c}{{\bf Occlusion 50\% (A)}} &
      \multicolumn{2}{c}{{\bf Corruption 70\% (A)}} &
      \multicolumn{2}{c}{{\bf Mixture Noise (Y)}} \\
      & {Accuracy} & {Time} & {Accuracy} & {Time} & {Accuracy} & {Time} & {Accuracy} & {Time} & {Accuracy} & {Time} \\
      \midrule
    RRC\_L1 \cite{Yang2013} & 97.14\% & 6.88s & 78.24\% & 12.41s & 59.57\% & 3.95s & \;\;\;{\bf 91.92\%} & 29.84s & 43.08\% & 14.58s   \\    
    RRC\_L2 \cite{Yang2013} & 97.37\% & 31.11s & 81.54\% & 10.52s  & 55.57\% & 1.96s & 88.38\% & 10.93s & 41.54\% & 9.65s  \\ \hline
    F-IRLS & \;\;\;{\bf 98.06\%}  & 1.32s & 83.96\% & 1.74s & 55.86\% & 0.39s &  87.92\% & 1.17s & 47.91\% & 1.70s  \\
    F-IRSC & 97.77\%  & 3.11s & 79.78\% & 1.62s & 60.43\% & 0.51s & 90.08\% & 2.16s & 45.05\% & 1.70s  \\
    F-IRNNLS & 98.00\% & 1.05s & 80.22\% & 1.52s & 62.29\% & 0.45s & 91.62\% & 2.25s & 45.27\% & 1.45s   \\ \hline
F-LR-IRLS & 96.74\% & 1.36s & \;\;\;{\bf 99.12\%}  & 2.81s & 71.71\% & 0.56s & n/a & n/a & 57.36\% & 9.38s \\
    F-LR-IRSC & 97.77\%  & 2.79s & 98.02\%  & 2.50s  & \;\;\;{\bf 74.57\%} & 0.68s & n/a & n/a & \;\;\;{\bf 66.81\%} & 7.96s   \\
    F-LR-IRNNLS & 97.83\% & 1.83s & 95.82\% & 2.41s & 74.43\% & 0.57s & n/a  & n/a & 63.08\% & 6.33s   \\ 
    \bottomrule
  \end{tabular}
}
\end{center}
\label{tb:our_vs_rrc}
\vspace{-0.5cm}
\end{table*}

\subsection{Weight Map Estimations}\label{sec:weight_map}

Figure~\ref{fig:weight_results} shows the estimated weight maps between RRC\_L1, RRC\_L2 and our F-LR-IRNNLS in experiments with occlusions. Black values (close to zero) represent detected outliers by the various methods. We observe that F-LR-IRNNLS detected the outlier objects more effectively than the other methods. Most of the black regions in the weight maps are concentrated on the occluded area. In particular, we observe in Figure~\ref{fig:weight_results} (first row) that for the F-LR-IRNNLS method small weights are only assigned to the occluded (baboon) region as desired. On the other hand, the weight maps of RRC\_L1 and RRC\_L2 are not as accurate since outliers were detected in important pixels of the face. The reason is that with these methods there is no spatial correlation constraint between the weights. Similar conclusions can be drawn from all other examples in Figure~\ref{fig:weight_results}. 

\begin{figure}[!t]
     \centering
	\includegraphics[scale=0.28]{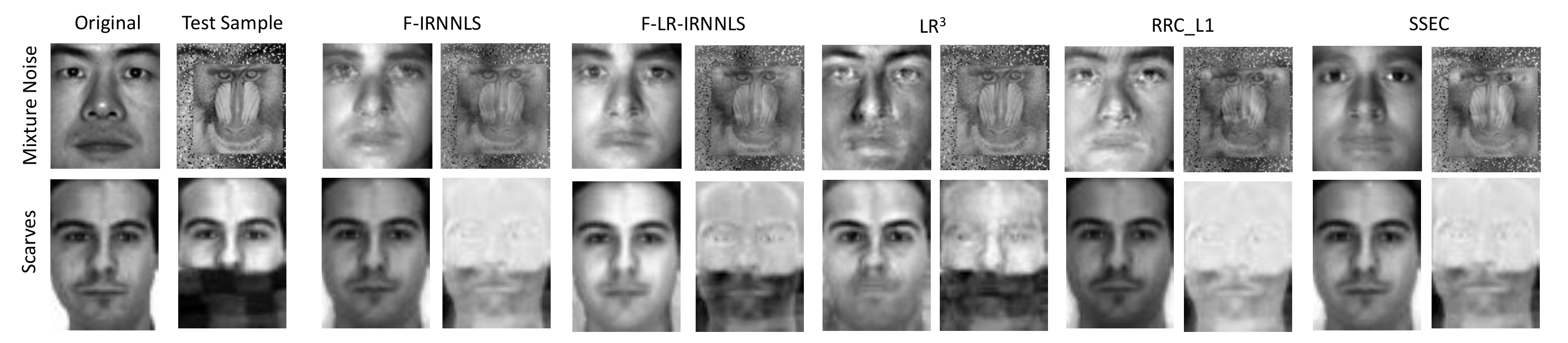}\\
     %\vspace{-0.3cm}
    \caption{Reconstruction results for various methods: The left image demonstrates the reconstructed face and the right image shows the estimated error for each of the methods tested.}
     %\vspace{0.0cm}
     \label{fig:qual_results}
\vspace{-0.3cm}
\end{figure}

%\begin{figure}[!t]
%\centering 
%\includegraphics[scale=0.34]{images/method_iterations_cropped.pdf}\\
%\caption{Reconstruction results of the error and low-rank images estimated during the ADMM iterations.} 
%\label{fig:qualitative_results_a}
%\vspace{-0.5cm}
%\end{figure}

\subsection{Face Reconstruction Results}

Figure~\ref{fig:qualitative_results_b} illustrates the face reconstruction results and the associated representation coefficients by the four methods. F-LR-IRNNLS and SSEC had the best reconstruction performance. The reconstructed face by $\text{LR}^3$ was poor mainly due to the choice of the regularizer for the representation coefficients ($\ell_2$ norm). Similar reconstruction performance for the $\text{LR}^3$ method was encountered in almost all of our conducted experiments. 

More reconstruction results for various methods are presented in Figure~\ref{fig:qual_results}. With mixture noise, our F-LR-IRNNLS achieved the best performance which demonstrates that our modeling was more effective in this case than the other methods. Face reconstruction was adequate for the case with scarves occlusion for all methods which validates the identification rates reported in Table~\ref{tb:disguise}.              

%Figure~\ref{fig:qualitative_results_a} shows the (weighted) error and low-rank estimations during the ADMM iterations between F-LR-IRNNLS and $\text{LR}^3$ methods. From the error images we observe that our two-step approach in \eqref{eq:nn_proximity} (weighted and low-rank projections) estimates the error accurately during the ADMM iterations. In particular, the final error and reconstructed face are more accurately estimated by F-LR-IRNNLS than by $\text{LR}^3$. 

\subsection{Time Performance between our method and RRC}

\begin{figure}[t]
\centering
\begin{subfigure}[b]{0.20\textwidth}\includegraphics[width=\textwidth]{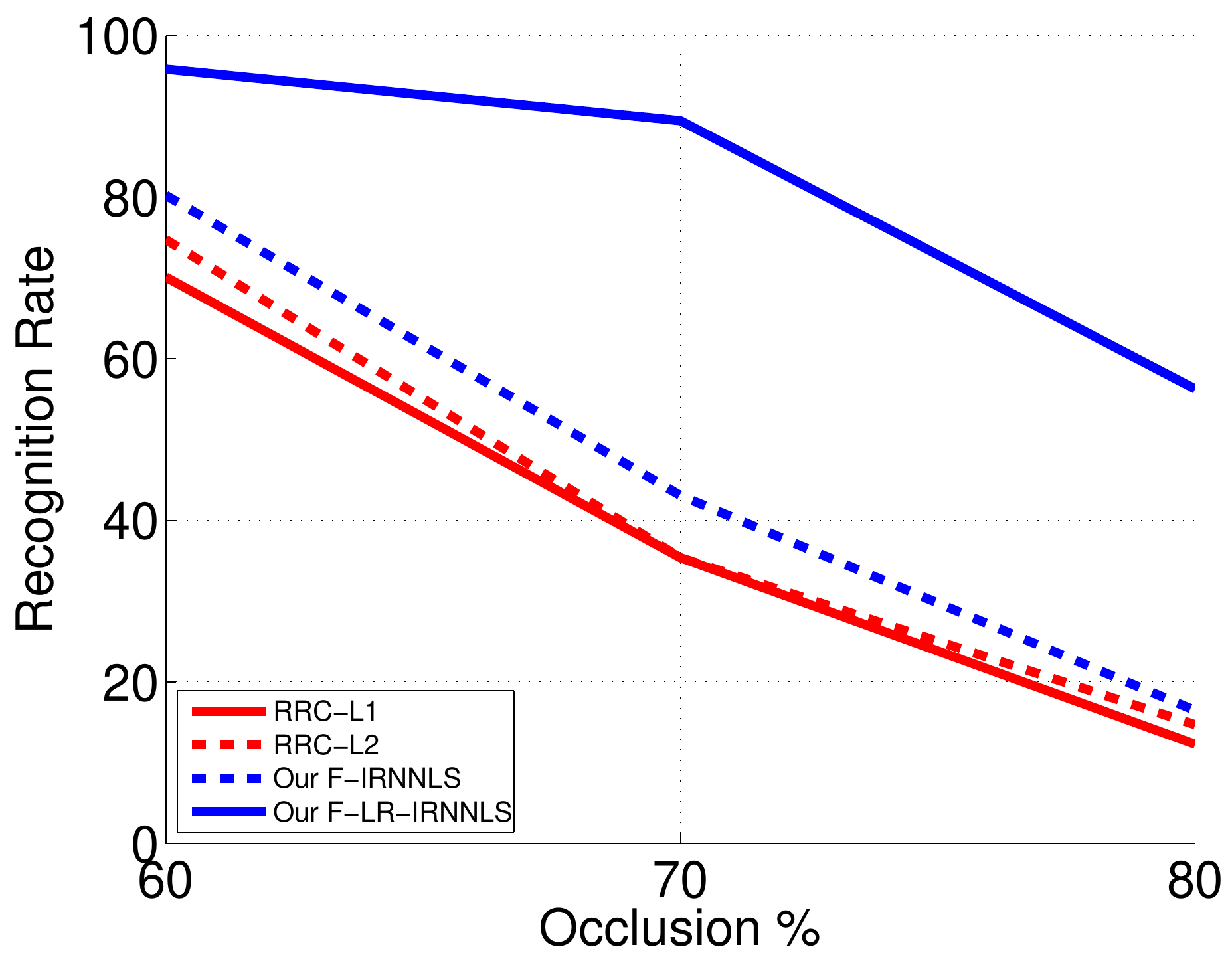}
\end{subfigure}%\hspace{0.1cm}
\begin{subfigure}[b]{0.20\textwidth}\includegraphics[width=\textwidth]{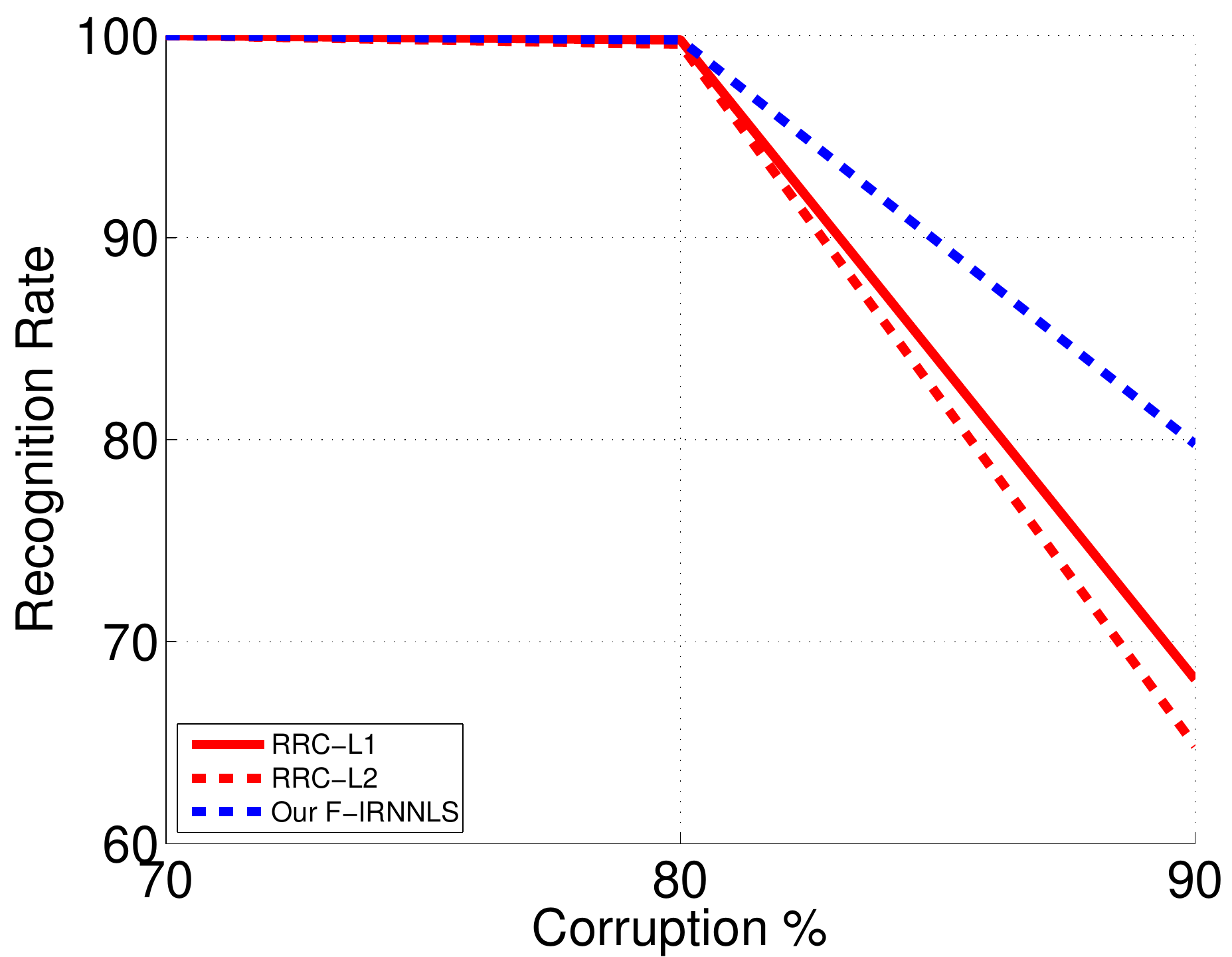}
\end{subfigure}\\%\hspace{0.1cm}
\begin{center}
\resizebox{!}{1.20cm}{
  \begin{tabular}{lSSSSSSSSSS}
    \toprule
    \multirow{2}{*}{Dataset} &
      \multicolumn{2}{c}{{\centering Yale (60\% occl.)}} &
      \multicolumn{2}{c}{{\centering Yale (90\% corr.)}} \\   
      & {Accuracy} & {Time} & {Accuracy} & {Time}  \\
      \midrule
  RRC\_L1 \cite{Yang2013} &70.11\%&9.16&68.13\%&8.75\\
RRC\_L2 \cite{Yang2013} &74.73\%&3.69&64.84\%&3.77\\
Our F-IRNNLS &80.22\%&{\bf \; 1.52}&{\bf \;\; 79.78}\%&{\bf \; 2.84}\\ 
Our F-LR-IRNNLS &{\bf \;\; 95.82}\%&2.41&n/a & n/a   \\ \hline
    \bottomrule
  \end{tabular}
}
\end{center}
\caption{Performance of RRC ($t = 25$ iterations) and our method ($t = 100$ iterations) under Block Occlusion and Pixel Corruption.} \label{tb:rrc_fastirls_time}
\vspace{-0.5cm}
\end{figure}

In this experiment we evaluate the identification rates in RRC \cite{Yang2013} for the case where the maximum reweighted iterations $t = 25$. In other words, we want to investigate the performance degradation of RRC by keeping its execution time similar to our method. In our method we kept $t = 100$. As shown in Figure~\ref{tb:rrc_fastirls_time}, we observe that the computational time for RRC is now more competitive (although still higher than our method). However, the identification rates dropped significantly in both pixel corruption and block occlusion cases for RRC with $t = 25$.

\subsection{Regularization of the Coefficients}

In Table~\ref{tb:our_vs_rrc} we report performance comparisons of our method with different regularizations of the representation coefficients. Our main take away from the results is that sparsity is overall slightly better than the two other regularizers (non-negative and $\ell_2$) in terms of identification rates. However, the non-negative regularizer provided a better balance between computational cost and identification rates.

Finally, there is significant difference in time performance between the RRC and our method regardless of the regularization of the coefficients. The efficiency of our method gives rise to robust face recognition systems for which computational time is a critical factor.

\section{Conclusions}\label{sec:conclusions}

In this work we proposed a method to describe contiguous errors effectively based on two characteristics. The first fits to the errors a distribution described by a tailored loss function. The second describes the error image as structural (low-rank). Our approach is computationally efficient due to the utilization of ADMM. The extensive experimental results support the claim that the proposed modeling of the error term can be beneficial and more robust than previous state-of-the-art methods to handle occlusions across a multitude of databases and in different scenarios. A special case of our algorithm leads to the robust representation problem which is used to solve cases with non-contiguous errors. We showed that our fast iterative algorithm was in some cases faster by an order of magnitude than the existing approaches. 

%\section{Acknowledgments}

%The paper has been supported by the Spanish Ministry of Economy and Competitiveness under project TIN2013-43880-R.

% Note that the IEEE does not put floats in the very first column
% - or typically anywhere on the first page for that matter. Also,
% in-text middle ("here") positioning is typically not used, but it
% is allowed and encouraged for Computer Society conferences (but
% not Computer Society journals). Most IEEE journals/conferences use
% top floats exclusively. 
% Note that, LaTeX2e, unlike IEEE journals/conferences, places
% footnotes above bottom floats. This can be corrected via the
% \fnbelowfloat command of the stfloats package.

\ifCLASSOPTIONcaptionsoff
  \newpage
\fi



%\bibliographystyle{IEEEtran}
%\FloatBarrier

\balance
\bibliography{facebib}



%\vfill


\end{document}